\pdfoutput=1

\documentclass[runningheads]{llncs}

\usepackage{amsmath}
\usepackage{graphicx}
\usepackage{xcolor}
\usepackage[utf8]{inputenc}
\usepackage{subfigure}

\usepackage{amsmath}

\begin{document}

\title{SegXAL: Explainable Active Learning for Semantic Segmentation in Driving Scene Scenarios}

\author{
Sriram Mandalika\inst{1} \and
Athira Nambiar\inst{1}}

\authorrunning{Sriram Mandalika et al.}

\institute{Department of Computational Intelligence,\\
Faculty of Engineering and Technology,\\
SRM Institute of Science and Technology\\
Kattankulathur, Tamil Nadu, 603203, India\\ 
\email{mc9991@srmist.edu.in, athiram@srmist.edu.in}}

\maketitle             
\begin{abstract} Most of the sophisticated AI models utilize huge amounts of annotated data and heavy training to achieve high-end performance. 
However, there are certain challenges 
that hinders the deployment of AI models ``in-the-wild" scenarios i.e. inefficient use of unlabeled data, lack of incorporation of human expertise and lack of interpretation of the results. To mitigate these challenges, we propose a novel Explainable Active Learning (XAL) model \textit{viz.} `\textbf{XAL-based semantic segmentation model ``SegXAL" }, that can (i) effectively utilize the unlabeled data, (ii) facilitate the ``Human-in-the-loop” paradigm and (iii) augment the model decisions in an interpretable way. In particular, we investigate the application of the SegXAL model for semantic segmentation in driving scene scenarios. The SegXAL model proposes the image regions that require labelling assistance from Oracle by dint of explainable AI (XAI) and uncertainty measures in a weakly-supervised manner. Specifically, we propose a novel Proximity-aware Explainable-AI (PAE) module and Entropy-based Uncertainty (EBU) module to get an Explainable Error Mask, which enables the machine teachers/human experts to provide intuitive reasoning behind the results and to solicit feedback to the AI system, via an active learning strategy. Such a mechanism bridges the semantic gap between man and machine through collaborative intelligence, where humans and AI actively enhance each other’s complementary strengths. A novel high-confidence sample selection technique based on the DICE similarity coefficient is also presented within the SegXAL framework. Extensive quantitative and qualitative analyses are carried out in the benchmarking Cityscape dataset. Results show the outperformance of our proposed SegXAL against other state-of-the-art models.

\end{abstract}

\section{Introduction}

Over the past decade, the world has witnessed an unprecedented technological revolution with the help of Artificial Intelligence (AI) towards accelerating automation, improving decision-making processes, and extracting insights from vast datasets. Despite these advancements, deep learning models commonly encounter substantial challenges while deploying in real-world or ``in-the-wild" settings, such as limitation of well-annotated data, contextual \& prior information and interpretability of the results~\cite{talaei2023deep}.

\textit{Annotation of new data points} is an expensive and laborious task, yet crucial for enriching training datasets with valuable information. In tasks like image semantic segmentation, manually labelling each pixel with its class label is arduous. Supervised algorithms provide efficient solutions for this task, whereas in unsupervised scenarios, automatic labelling poses a significant challenge for machines. Furthermore, \textit{integrating prior and contextual information} can significantly enhance AI model performance, especially in  high-risk scenarios e.g. medical and defence. Domain experts can contribute valuable knowledge to AI systems in such situations, enabling a ``Human-in-the-loop" paradigm for more rational analysis and informative results. However, most existing AI systems lack mechanisms to incorporate \textit{additional human-collected information or domain expertise}. In real-world scenarios, the inverse situation also exists, wherein the operators often have to rely on visual inspection to make decisions due to the \textit{lack of explainability in machine decisions}. Despite the advancements in deep neural networks, the integration of AI tools in various fields is hindered by the opacity of these ``black-box" models, which fail to provide explanations for their actions. All of these scenarios highlight the semantic gap between human and machine analysis, emphasizing the need for human involvement in decision-making as well as the development of Explainable AI tools towards better interpretability of the model.

To mitigate the aforementioned challenges, we propose a novel Explainable Active Learning (XAL) model that 
combines domain expert assistance and explainable AI (XAI) support within the active learning (AL) paradigm.

In particular, we propose a novel \textbf{XAL based semantic segmentation model ``SegXAL"} for the driving scene scenarios. \textit{Active learning} facilitates effective training set by iteratively curating the most informative unlabeled data for annotation with the help of human intervention (oracle)
accentuating the ``human-in-the-loop" paradigm \cite{lewis1994heterogeneous},\cite{margatina2021active}. This “domain expert teaching”  emphasizes productivity and enhances trust in AI systems, especially in low-resource as well as high-risk scenarios. Similarly, the \textit{explainability} aspect of the SegXAL model enables the “machine teachers” (human experts) to obtain intuitive reasoning behind the results and to give solicit feedback to the system\cite{arrieta2020explainable}. This is inspired by the rationale that humans' cognizance leverages causal and interpretable information to make decisions\cite{rottman2014reasoning}, \cite{yang2022psychological}. Both of these AL and XAI notions within the SegXAL model bridge the semantic gap between man and machine through collaborative intelligence, wherein humans and AI actively enhance each other’s complementary strengths.

The key component of the SegXAL framework is the Explainable Error Mask (EEM) module that provides intuitive reasoning as well as uncertainty measures for the sample selection. The EEM module internally contains two components viz. Entropy-based Uncertainty (EBU) module and Proximity-aware Explainability (PAE) module. Following popular active learning approaches, the EBU module utilizes uncertainty or disagreement in the unlabeled data to identify the most uncertain and informative samples for annotation by the oracle \cite{Xie_2020_ACCV} \cite{siddiqui2020viewal} \cite{lenczner2022dial}. Whereas, the PAE module acts as an interpretable proximity approximator that prioritizes the relevant nearby class information, leveraging depth estimation technique and explainable AI. \textcolor{black}{In particular, two advanced AI models viz. MiDaS \cite{ranftl2020towards} and DINOv2 \cite{OquabDINO2023} are used as the instances of depth estimation. Refering to the XAI technique, 
we leverage Gradient-weighted Class Activation Mapping (GradCAM) \cite{selvaraju2017grad}, which interprets and visualizes the regions of an input image that are crucial for the network's prediction of a specific class.}

Thus, the PAE module along with the EBU module provides the Explainable-Error Mask (EEM) with both informativeness and explainability, thereby facilitating meaningful annotation from the oracle. Two modes of oracle annotations are presented in this work: The first mode is via \textcolor{black}{\textbf{Machine annotated pseudolabels}}, wherein the machine itself does an automatic pixel annotation. The second mode is via \textbf{Manual annotation}, wherein the human annotator labels the region relevant to the object based on the candidate prompts. The major contributions of the paper are as follows:
\vspace{-.3cm}
\begin{itemize}
    \item Proposal of a \textbf{‘XAL based semantic segmentation model ``SegXAL" for the driving scene scenarios’}, \textcolor{black}{which is the first Explainable Active Learning (XAL) framework in semantic segmentation.}
    \item Development of a novel \textbf{Explainable Error Mask (EEM)}, fusing proximity-aware explainability (PAE) and entropy-based uncertainty (EBU) measures, thereby enhancing the efficiency of oracle annotation.
    \item \textcolor{black}{Proposal of two manual annotations schemes within the Active learning framework viz. \textbf{Manual-M and Manual-D}, leveraging MiDaS and DINOv2-based explainable error masks, respectively.}
    \item Proposal of a novel \textbf{high-confidence sample selection technique based on DICE} similarity coefficient.
    
    \item Extensive experimental analysis, ablation studies and state-of-the-art comparative analysis in benchmarking Cityscapes dataset.

\end{itemize}
\vspace{-.3cm}
The rest of the paper is organized as follows: The related works are described in Section~\ref{sec:related}. The proposed SegXAL  active learning framework is presented in Section~\ref{sec:methodology}. The experimental setup and the results are discussed in detail in Section~\ref{sec:setup} and Section~\ref{sec:results} respectively. Finally, the summary of the paper and some future plans are enumerated in Section~\ref{sec:conclusion}.

\vspace{-.3cm}
\section{Related Works}
\label{sec:related}
\vspace{-.2cm}
\noindent \textbf{Explainable AI}:
Explainable Artificial Intelligence (XAI) is an emerging area of research in machine learning \cite{yang2022psychological}. XAI techniques make AI models more interpretable by humans by divulging the hidden ``black-box" and providing insights into how the model arrives at a particular decision. Some of the recent research works have been investigating XAI in such cutting-edge areas, e.g. medical domain to find out the feature importance~\cite{zuallaert2018splicerover} and to visualize the biologically relevant information~\cite{rajpurkar2020appendixnet}. Some XAI models were developed for remote sensing and satellite applications,~\cite{stomberg2021jungle} to analyze synthetic aperture sonar (SAS) data and for Explainable Machine Learning in Satellite Imagery, respectively. The application of XAI approaches in driving scene scenarios is also reported in the recent literature bestowing ideas towards comprehensible and trustworthy autonomous driving technologies~\cite{atakishiyev2021towards}.\\

\noindent \textbf{Active Learning}: Active Learning (AL) entails the training process of a learning algorithm through an iterative collaboration
with a human oracle~\cite{settles2009active}. AL involves selecting the most relevant data samples from a pool of unlabeled data based on uncertainty, representativeness, or diversity scores computed directly with the model \cite{Xie_2020_ACCV},\cite{siddiqui2020viewal}. To this end, some popular approaches to obtain confidence, margin and uncertainty measures  are via entropy~\cite{settles2008analysis}, Softmax probabilities~\cite{wang2016cost}, Monte Carlo dropout~\cite{gal2017deep} and Ensemble methods~\cite{beluch2018power}. Such AL models have been widely applied in various vision applications,
such as medical scenarios~\cite{liebgott2016active}, satellite imagery analysis~\cite{goupilleau2021active} etc. The necessity for AL frameworks for autonomous driving scenarios is reported in~\cite{nvidia-autonomous}, mentioning that 'vehicles need 11 billion miles of driving (500 years of nonstop driving with a fleet of 100 cars) to perform just 20 per cent better than a human.' Motivated by this notion, some recent AL works on driving scenes were reported in the literature ~\cite{schmidt2020advanced}.\\

\noindent \textbf{AL for semantic segmentation}: There are AL methods specially designed for semantic segmentation that work at image, region or pixel levels~\cite{Xie_2020_ACCV},~\cite{siddiqui2020viewal}. The Variational Adversarial Active Learning (VAAL) approach employs adversarial learning to determine whether the latent space signifies labelled or unlabeled data~\cite{sinha2019variational}. The work Difficulty-awarE Active Learning (DEAL)~\cite{Xie_2020_ACCV} incorporates the semantic difficulty to measure the informativeness and select samples at the image level. Another work `ViewAL'~\cite{siddiqui2020viewal} leverages inconsistencies in model predictions across view-points to measure the uncertainty of super-pixels. Yet another work S4AL~\cite{rangnekar2023semantic} utilizes pseudo labels generated with a teacher-student framework to identify image regions that help disambiguate confused classes.\\

\vspace{-.2cm}

Contrary to the aforementioned AL approaches that measure uncertainty/ informativeness, our proposed SegXAL additionally augments the notion of explainability in the model. In particular, the PAE module in our proposed SegXAL model imparts contextual and proximity-aware explainability to the oracle to prioritize the annotation of nearby objects, which are pivotal in autonomous driving scenarios. \textcolor{black}{This kind of explainable active learning (XAL) in semantic segmentation is proposed for the first time, to the best of our knowledge.} Further, the significance of pixel-level and object-level annotation by the oracle (Machine annotator vs. Human annotator) is also investigated in our proposal.

\vspace{-.4cm}
\section{Methodology: SegXAL - Explainable Active Learning for semantic segmentation}
\label{sec:methodology}
\vspace{-.2cm}
The Active Learning (AL) protocol ensures that by intelligently selecting instances for labelling, a learning algorithm can achieve good performance with significantly less training data. Formally, it can be expressed as follows: Let $(x^l, y^l)$ be an annotated sample from the original labelled dataset $D^L$ and $x^u$  represent an unannotaed sample from a significantly larger pool of unlabeled data, $D^U$. The goal of AL is to  iteratively query a subset $D^S$, that contains the most informative $n$ samples ${x^{u}_{1}, x^{u}_{2}, ..., x^{u}_{n}}$ from $D^U$ in an iterative manner, given $n$ is the fixed labelling budget.\nocite{*}

In this work, we present a novel Explainable Active Learning paradigm for semantic segmentation  (SegXAL) in driving scene imagery. Refer to Fig. \ref{fig:SegXAL_architecture} for the overall architecture of the SegXAL framework. It contains training of the model, prediction of semantic maps, ``Explainable Error Mask" (EEM) computation, annotation, selection mechanism and retraining steps. Each of these steps is explained in detail in the forthcoming subsections:

\begin{figure}[t!]
\centering
\includegraphics[width=1.1\textwidth]{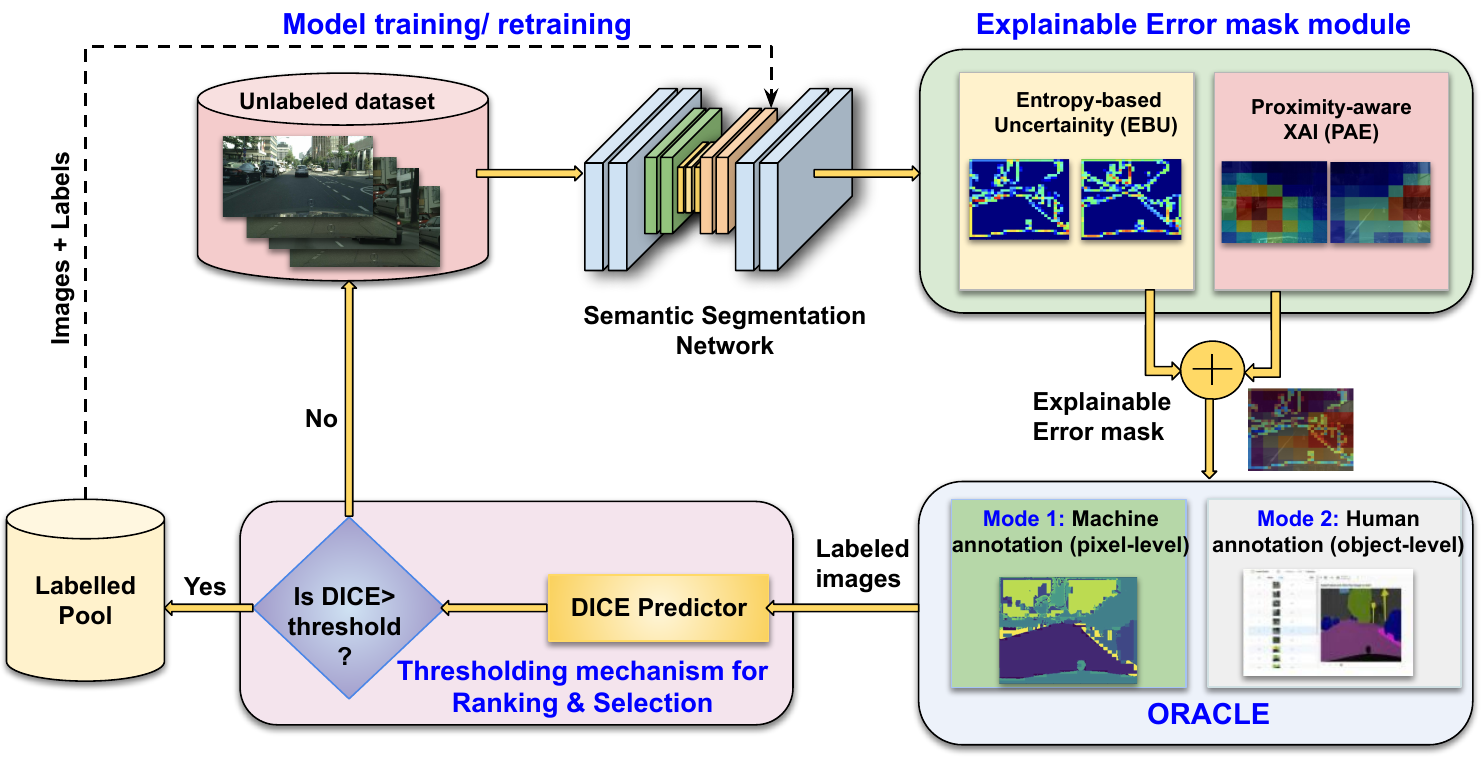}
\caption{Visual representation of Explainable Active Learning for
semantic segmentation (\textbf{SegXAL}) framework. The framework starts with an initial segmentation of unlabeled data, leveraging pre-trained semantic segmentation deep neural network (e.g. U-net). Further, the Explainable Error Mask (EEM) module computes the uncertainty measure and proximity-aware XAI mask. Based on this EEM output, machine/human expert (oracle) makes intuitive labelling feedback to the system. Further, based on the Dice predictor-based query ranking mechanism, reannotated data are used for labeled pool update and model retraining. }
\vspace{-.5cm}
\label{fig:SegXAL_architecture}
\end{figure}

\vspace{-.6cm}
\subsection{\textbf{Step 1: Semantic Segmentation - Training \& Prediction}}
\vspace{-.1cm}
We leverage U-Net~\cite{ronneberger2015u} as the semantic segmentation network for the model training. Typically, any segmentation model such as FCN \cite{long2015fully} or DeepLab~\cite{chen2017deeplab}, among others, could also be utilized. U-Net is employed in this pilot study, due to its ability for the precise localization of objects while maintaining a high level of contextual information as well as lower memory consumption. The U-Net model embodies an encoder-decoder framework. The encoder is responsible for the initial feature extraction and dimensionality reduction, by utilizing successive convolutional and pooling layers followed by nonlinear activation functions (ReLUs) and batch normalization. Whereas, the decoder works on reconstructing the feature map to the original image size for detailed segmentation using transposed convolutions (or deconvolutions). It also incorporates skip connections, that concatenate feature maps from the contracting path to preserve the high-resolution details that are crucial for accurate segmentation. 

In this initial step, a small randomly selected subset of the labelled dataset $D_L$ will be used to train a semantic segmentation network. 
Following the widely adopted protocol, we randomly sample 10\%  of the data as labelled data from the train set as our labelled data pool\footnote{(Ablation studies are carried out by varying the splits of labelled data pool i.e. 10\%, 15\%, 20\%, 25\%, 30\%, 35\%, 40\%)}. After training the network on $D^L$, the model performance is evaluated on unlabeled dataset $D^U$. AL approach strives to forecast which samples from this unlabeled segment of dataset, are most likely to provide the most informative insights, given the current state of the network. To this end, a novel \textbf{Explainable Error Mask (EEM)} module is proposed.

\vspace{-.5cm}
\subsection{\textbf{Step 2: Explainable Error mask Module}}
\label{EEM}
The Explainable Error Mask (EEM) module is the key component of our SegXAL framework. In contrast to the vanilla Active learning models that provide uncertainty/ representativeness insights for the annotation, this novel EEM module presents an explainable error mask for the interactive annotation by the oracle. It consists of the following components: \textit{i) Entropy-based Uncertainty (EBU)}, \textit{ii) Proximity-aware XAI (PAE)} and \textit{iii) fusion of PAE and EBU}.

\vspace{-.5cm}
\subsubsection{i) Entropy-based Uncertainty (EBU) module:\\}
One of the most important postulations in active learning strategy is to guide the user towards the most relevant areas to annotate, to fix errors. To this end, some standard uncertainty measuring techniques such as entropy~\cite{shannon1948mathematical}, or ODIN~\cite{liang2017principled} are exploited in the literature. Following many of the popular AL pipelines, our EBU module leverages
entropy metric to measure the uncertainty/ disagreement for the unlabeled data, to obtain the most uncertain data which is informative and worthful ones to be annotated by the oracle.

Entropy is a measure of uncertainty or information content in a probability distribution~\cite{shannon1948mathematical}. In the context of image segmentation, it is commonly used to quantify the uncertainty of pixel-wise predictions across different classes within a batch of segmented images. Let us denote a batch of segmented images as $X$ with dimensions $[B,C,H,W]$, where $B$ is the batch size, $C$ is the number of classes, $H$ is the height and $W$ is the width of images. Each image in the batch consists of pixel-wise predictions across $C$ classes. The entropy $H(x_{i,j})$ for each pixel $x_{i,j}$ can be calculated as:

\vspace{-.5cm}
\begin{equation}
H(x_{i,j}) = -\sum_{c=1}^{C} P(c | x_{i,j}) \log_2(P(c | x_{i,j}))
\end{equation} 
where $P(c|x_i)$ represents the probability that pixel $x_{i,j}$ belongs to class $c$.
Higher entropy values indicate greater uncertainty or ambiguity in the predictions, implying lower confidence in the model's predictions. Conversely, lower entropy values signify higher confidence or clarity in the predictions.

\vspace{-.6cm}
\subsubsection{ii) Proximity-aware Explainable-AI (PAE) module:\\}

\begin{figure}[t!]
\centering
\includegraphics[width=\textwidth]{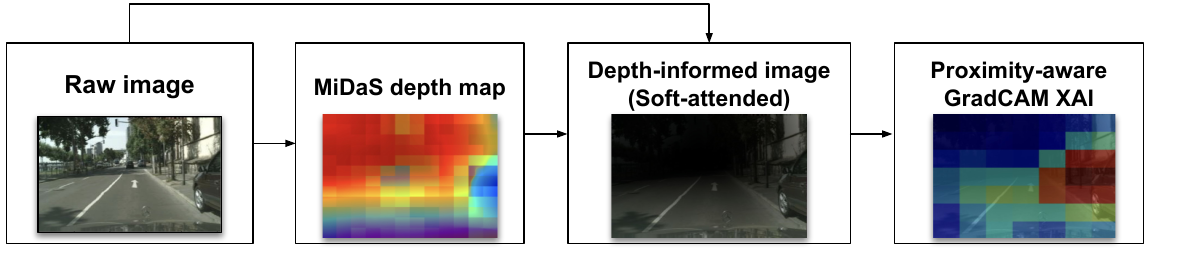}
\caption{Proximity-aware Explainable-AI (PAE) Module using MiDaS depth estimation technique. Analogous to MiDaS, DINOv2 depth map is also investigated in this paper.}
\vspace{-.6cm}
\label{fig:PAE}
\end{figure}

The high entropy pixels generated by the Entropy-based Uncertainty (EBU) module can be spread across the entire image, making it challenging from an Oracle perspective to determine where to prioritize attention. Consequently, this may lead to missing out of some of the vital regions to be annotated first. For instance, in driving scene imagery with high entropy scores in the sky, vegetation, and vehicles, annotation priority should be given to nearby classes i.e. vehicles, considering safety concerns. We hypothesise that such a proximity awareness can improve the oracle annotation. In addition, uncertainty techniques often lack human interpretability, hindering an intuitive understanding of why certain regions are crucial for annotation.

Based on the aforesaid rationale, we propose a novel \textbf{Proximity-aware Explainable-AI (PAE)} module to mitigate the priority and interpretability concerns. The PAE module is capable of focusing on the key objects and regions of interest in the proximity regions with the help of an explainability heatmap. The working pipeline of our proposed PAE module is depicted in Fig.~\ref{fig:PAE}.  \textcolor{black}{Either MiDaS or DINOv2 model is leveraged to obtain the given image's relative depth map. MiDaS~\cite{ranftl2020towards} is a robust monocular depth estimation technique that employs mixed-dataset training to create a robust and generalizable depth estimation model. Whereas, DINOv2~\cite{OquabDINO2023} is a self-supervised vision transformer model that uses a teacher-student architecture to provide object-level feature extraction. Both of the models are capable of providing monocular depth map outputs.} By integrating the MiDaS/DINO-v2 patchwise depth map with the raw image using a thresholding mechanism, the proximity coverage will be estimated. This results in a depth-informed or soft attention image as shown in Fig.~\ref{fig:PAE}. Note that the threshold for generating a depth-informed image varies with each image based on the proximity of the nearest objects. Upon this image, a Gradient-weighted Class Activation Mapping (GradCAM)~\cite{selvaraju2017grad} explainability map is applied to visualize the important objects and regions. GradCAM is a technique for visualizing CNN decisions, highlighting regions crucial for predictions. The mathematical equation for GradCAM activation at spatial position \((i,j)\) for class \(c\) i.e. $Grad-CAM^c_{i,j}$ can be summarized as:
\vspace{-.5cm}
\begin{equation}
\text{GradCAM}^c_{i,j} = \text{ReLU}\left(\sum_k \frac{1}{Z} \sum_i \sum_j \frac{\partial y^c}{\partial f_k(i,j)} \cdot f_k(i,j)\right) , 
\end{equation}
\vspace{-.1cm}
where, \( y^c \) is the output score for class \(c\) before softmax,  \( f_k(i,j) \) is the activation value of the \(k^{th}\) feature map at spatial position \((i,j)\) and  \( Z \) is the normalization constant, typically sum of positive gradients. By applying the GradCAM upon the depth-informed image, we obtain the proximity-aware GradCAM explainability map i.e. $ProxGradCAM^c_{i,j}$, which prioritizes the object class information which is relevant in the proximity region. 

\vspace{-.5cm}
\subsubsection{iii) \textcolor{black}{Fusion of PAE and EBU modules:\\}}
\textcolor{black}{The PAE heatmap $ProxGradCAM^c_{i,j}$ is further fused with EBU uncertainty heatmap $H(x_{i,j})$, to obtain the Explainable Error mask $EEM_{i,j}$. Formally, }
\vspace{-.2cm}
\begin{equation}
\textcolor{black}{EEM_{i,j} = \alpha \cdot ProxGradCAM^c_{i,j} + \beta \cdot \text{}H(x_{i,j})}
\label{eq:EEM}
\end{equation}
\vspace{-.1cm}
\textcolor{black}{where $\alpha$ and $\beta$ are the weights for the 
$ProxGrad-CAM^c_{i,j}$ and $H(x_{i,j})$, respectively. Albeit we used equal contribution for the weights in this work, it can be made learnable.}

\vspace{-.8cm}
\subsection{\textbf{Step 3: Oracle for annotation}}
\vspace{-.2cm}
Next, we acquire labels for the superpixels/Region of Interest (ROI) selected by EEM module, with the help of oracle. In particular, two modes of oracle annotations are envisaged in this work: machine and human oracle. In the former mode (Machine oracle), automatic pixel annotations are simulated by the machine itself. We term these annotations as `\textbf{pseudolabels}'. In the latter mode (Human oracle),  the reannotations are carried out manually by a domain expert. By keeping the interpretable information of the potential error map obtained from EEM as a reference, the annotation process is carried out using tools like Label Studio\footnote{Label studio: https://labelstud.io/}. Specifically, two manual annotation schemes are devised within the Active learning framework viz. \textbf{Manual-M} and \textbf{Manual-D}, leveraging MiDaS and DINOv2-based explainable error masks, respectively.

\vspace{-.6cm}
\begin{figure}[h!]
\scriptsize
  \centering
\subfigure[Raw input]{\includegraphics[width=2.35cm]{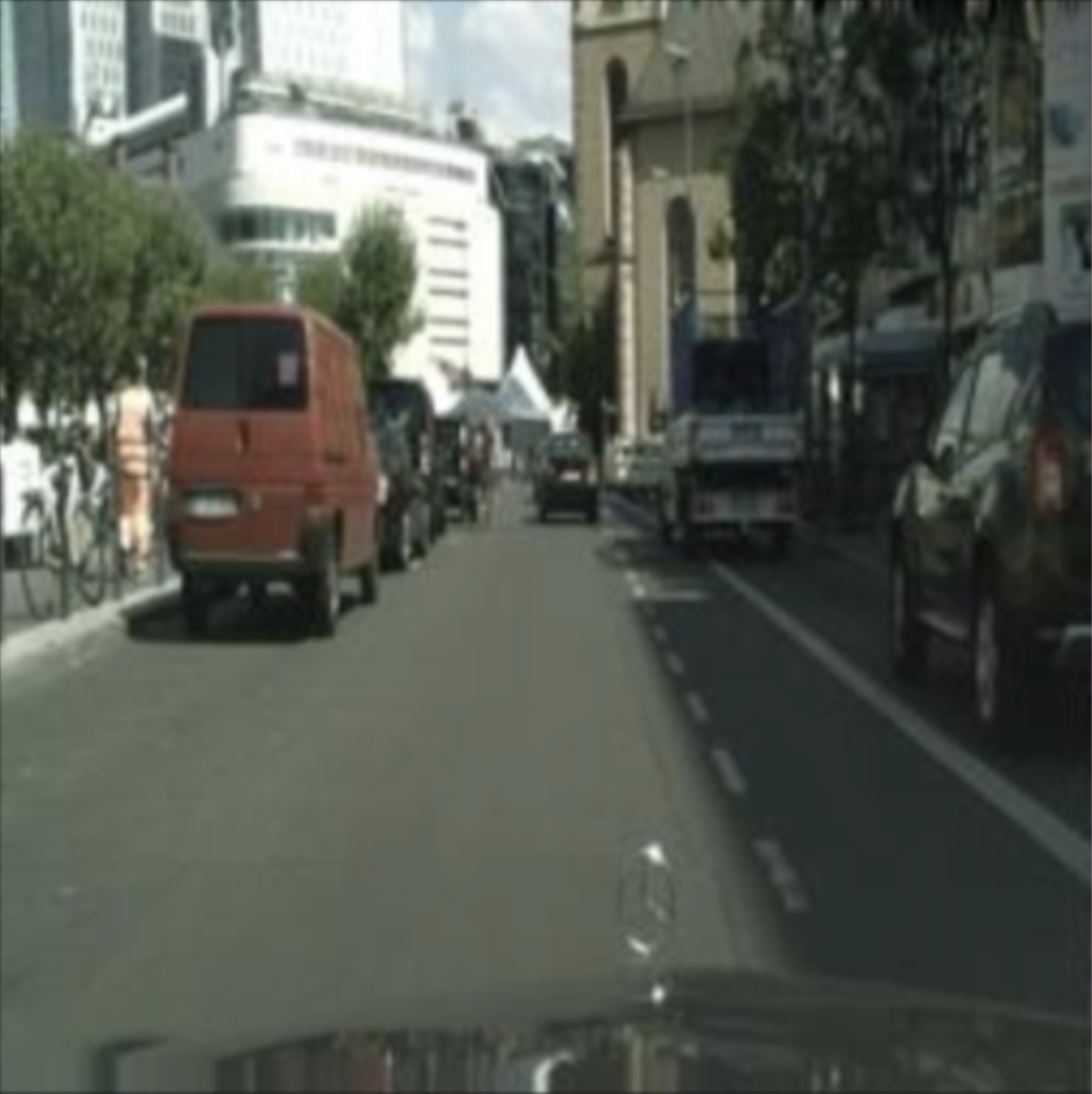}}
\subfigure[Initial output]{\includegraphics[width=2.35cm]{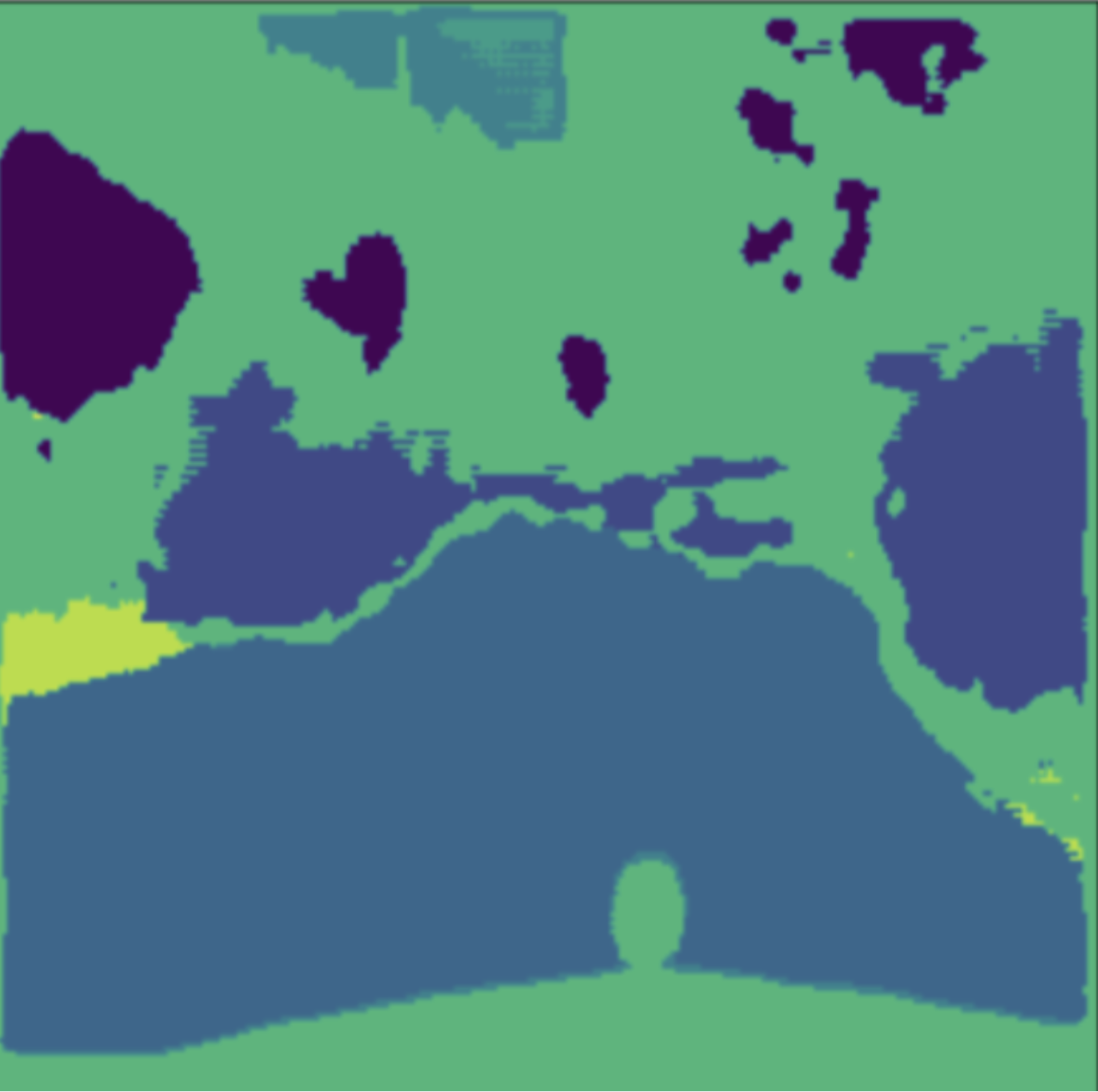}} 
\subfigure[EEM output]{\includegraphics[width=2.35cm]{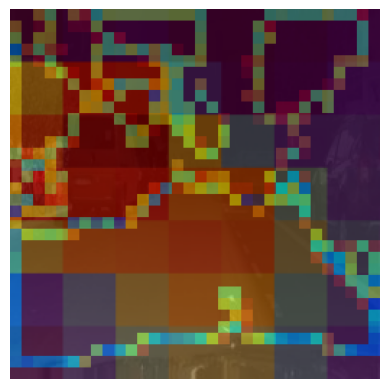}} 
\subfigure[Candidate prompt]
{\includegraphics[width=2.35cm]{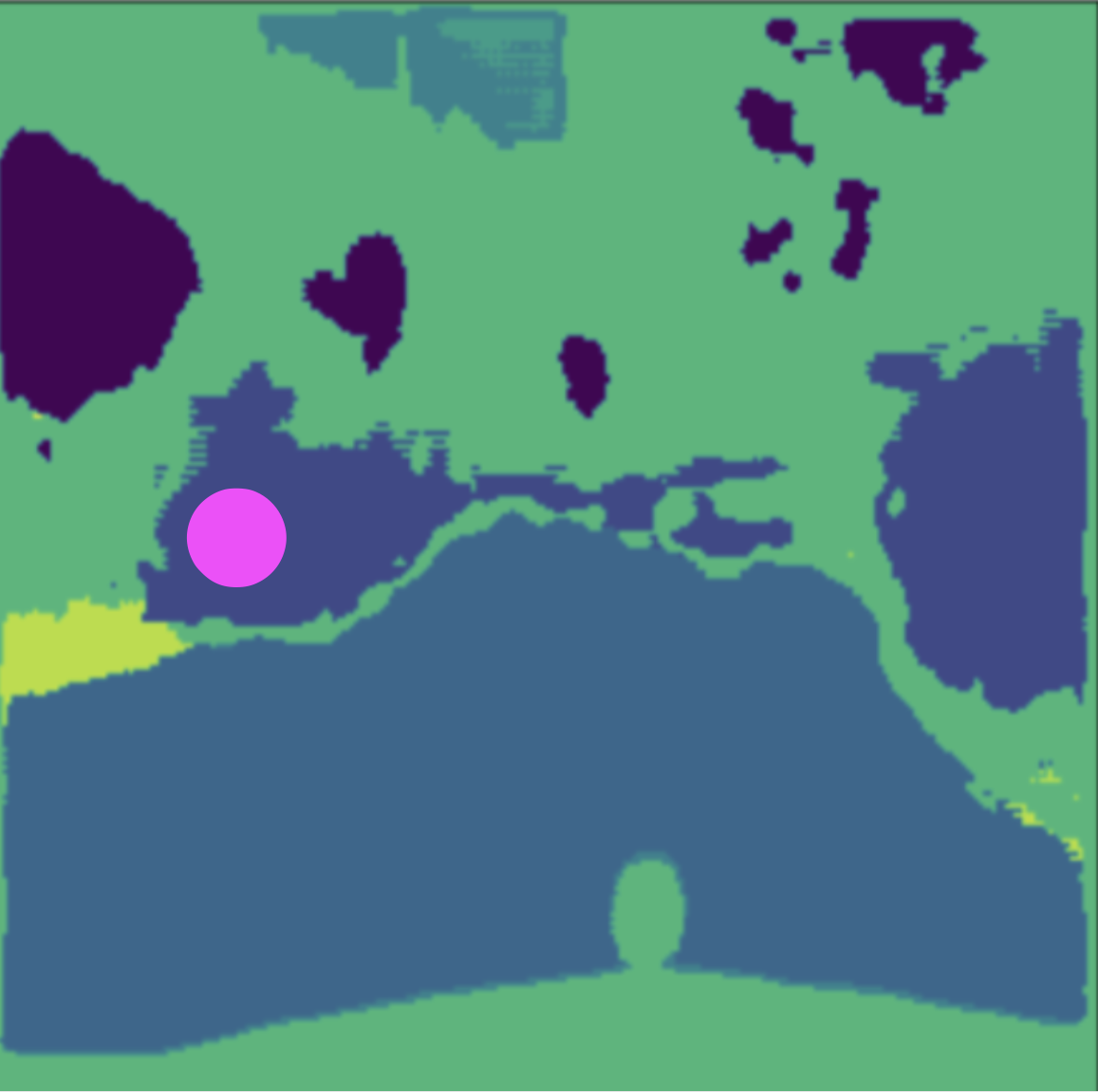}} 
\subfigure[Manual annotation]
{\includegraphics[width=2.35cm]{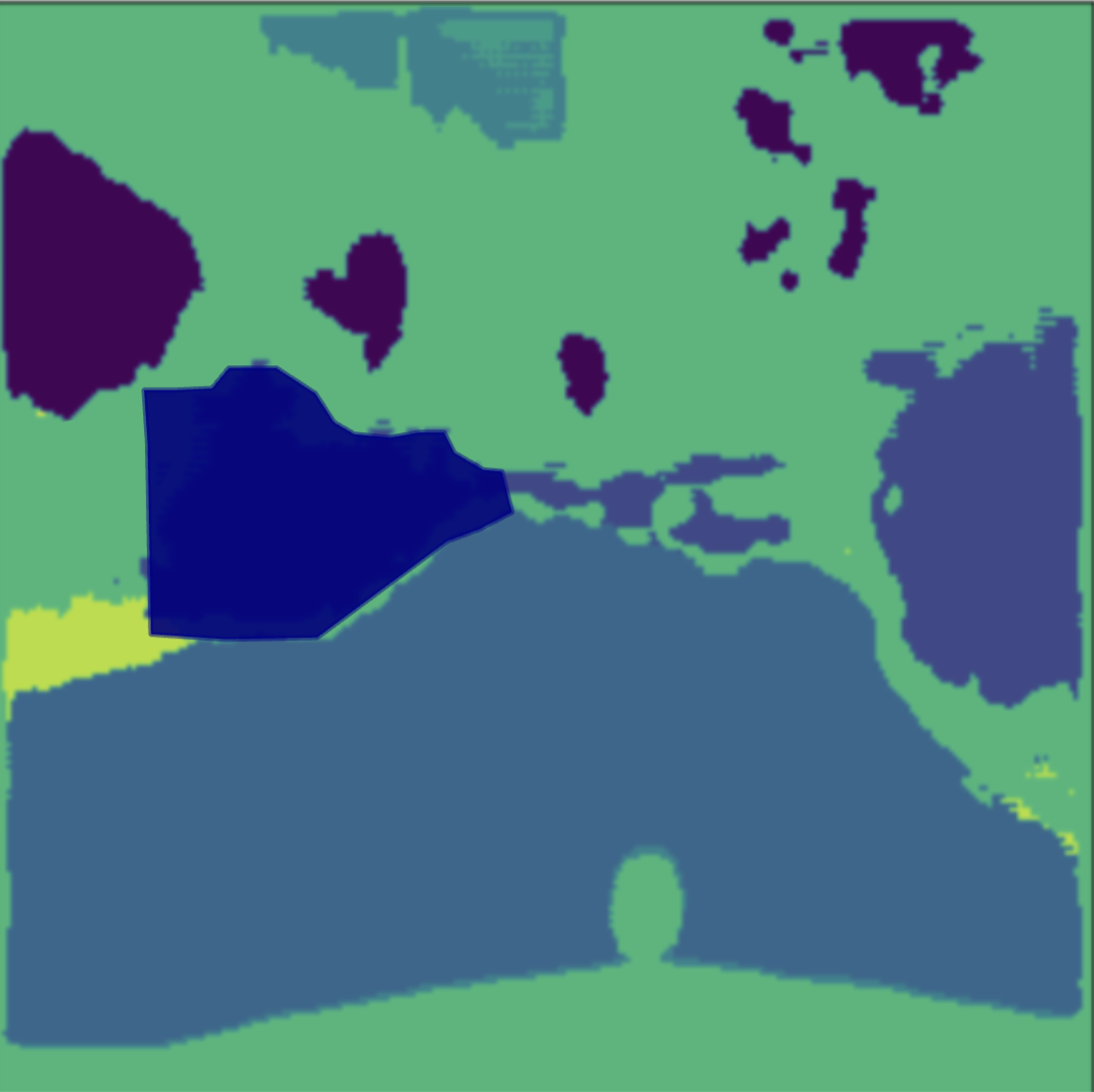}}

\vspace{-.2cm}
\caption{\small {\textcolor{black}{Oracle's Reannotation workflow. The magenta point shown in ~\ref{fig:revised-visuals}(d) is the EEM output prompt corresponding to the relevant object candidate to be annotated.}}}

\vspace{-.7cm}
\label{fig:revised-visuals}
\end{figure}

\textcolor{black}{Fig \ref{fig:revised-visuals} depicts a sample human oracle-based reannotation workflow. Based on the initial segmentation mask output from the raw image as shown in Fig. \ref{fig:revised-visuals}(b), EEM produces the output $EEM_{i,j}$ (Refer Fig. \ref{fig:revised-visuals}(c)). Further, based on the object candidate prompt 
as shown \ref{fig:revised-visuals}(d), the human annotator corrects the miss-segmented image regions by providing object-level annotation (Fig. \ref{fig:revised-visuals}(e)). These newly reannotated segmentation masks will be further fed into the sample selection module towards the next iteration of the AL loop. }

\vspace{-.5cm}
\subsection{\textbf{Step 4: Thresholding Mechanism for Sample Selection }}
After the oracle, the labeled images are fed into the Ranking \& Selection module. Analogous to the high-confidence sample selection techniques as in~\cite{wang2016cost}, we use a novel thresholding mechanism to select high-confidence samples to be incorporated into the labeled data pool. In particular, a standard evaluation metric i.e. `DICE predictor' is utilized to compute the quantitative measure of performance of the segmented images. Mathematically, DICE computation can be written as:
\vspace{-.2cm}
\begin{equation}
\text{DICE} = \frac{2 \times |A \cap B|}{|A| + |B|}
\end{equation}
where \( A \) represents the segmented image and \( B \) denotes the reannotated pseudo labels/ human annotations, within each AL cycle. 

This \textbf{DICE predictor-based sample selection strategy} is devised based on the assumption that \textit{``in every AL cycle, the oracle contributes a significant amount of annotation to improve the quality of semantic segmentation"}. Based on this intuition, we postulate that whenever the similarity between the  segmented image and the reannotated image becomes high, a convergence is achieved in the segmentation result. In other words, even after a significant amount of contribution from the oracle, the segmentation result does not improve further, which can be observed as an increase in the DICE similarity coefficient. To guarantee the reliability of high-confidence sample selection, at the end of each iteration, this DICE value is compared against a predefined threshold $\theta$. If the DICE score is above $\theta$, select it and add to the labeled pool and clear it from the unlabeled set; otherwise, feed it back to the unlabeled dataset placed in the unlabelled pool for potential future iterations.

\vspace{-.5cm}
\subsection{\textbf{Step 5: Iterative Active Loop for Semantic Segmentation Improvement}}

After the Ranking \& Selection module, high-confidence segmentation images are  added to the labelled data pool $D^L$, as shown in Fig. \ref{fig:SegXAL_architecture}. Based on this updated dataset, the semantic segmentation model retraining will be carried out. This concludes a complete active learning cycle. Further, a new AL cycle will start based on the updated model weights and the unlabeled dataset $D^U$. All the series of steps - \textit{Semantic map prediction from unlabelled data, EEM computation, Annotation, Ranking \& Selection and Retraining} - are repeated until the labelling budget is reached or all the data is labelled. This iterative AL cycle optimally selects the most informative samples via EEM information and Oracle annotation, enhancing model performance with minimal labelling costs.

\vspace{-.5cm}
\section{Experimental Setup}
\label{sec:setup}
\noindent \textbf{Dataset:} We evaluate our proposed SegXAL framework on the Cityscapes dataset for semantic segmentation~\cite{cordts2016cityscapes}. Cityscape is a large-scale benchmark for urban street scene understanding, at 
$1024 \times 2048$ pixel resolution with 30 classes including road, car, pedestrian, bicycle, traffic sign, and more.  The dataset is divided into three subsets: \textit{train} (2975 images), \textit{validation} (300 images), and \textit{test} (500 images). We follow the widely adopted protocol for the dataset - we sample 40\% of the data from the trainset as our labelled data pool  $D^L$ for initial training then iteratively query 5\% new data from the remaining training set, which is used as the unlabeled data pool  $D^U$. Considering samples in the street scenes have high similarities, we first randomly choose a subset  $D^S$ from the entire pool of  $D^U$, then query $n$ samples from the subset.\\
\noindent \textbf{Evaluation protocol:}
We evaluate our proposed SegXAL model using the standard segmentation evaluation metrics i.e. Intersection over Union (\textbf{IoU}) and DICE coefficient. To assess the accuracy of pixel-wise classification, the standard evaluation metric IoU (Intersection over Union) score is utilized. IoU is computed as the ratio of the intersection and union of the ground truth mask and the predicted mask for each class. Further, the DICE similarity coefficient is utilized for ranking \& selection of samples, as described in Section 3.4. It provides a balanced measure of segmentation accuracy, especially in cases of class imbalance, and hence is used for our sample selection strategy.\\
\noindent \textbf{Implementation details:}
The images with a dimension of \(256 \times 512\) are normalized using the RGB mean and standard deviation of ImageNet before passing to the network. Our baseline UNet model was evaluated using a stratified K-fold cross-validation approach to ensure robustness and generalizability. The network is trained using a Stochastic Gradient Descent (SGD) optimizer with the following hyper-parameters: $\beta$1 = 0.9, $\beta$2 = 0.999, batch size = 16, initial learning rate = 0.0001. The batch size used is 16 images. For all methods and the upper bound method with the full training data, we train 100 epochs with an unweighted cross-entropy loss function. The proposed method is implemented using the PyTorch framework.  The implementation was done in a machine with NVIDIA DGX A100 GPU with 24GB RAM and takes around 8 hours to train the model.

\vspace{-.5cm}
\section{Experimental Results}
\vspace{-.3cm}
\label{sec:results}
\subsection{Evaluation Results}
To verify the effectiveness of our proposed SegXAL framework, various quantitative and qualitative analyses are carried out in the Cityscape dataset.  The mean Intersection over Union (mIoU) at each AL stage i.e. 10\%, 15\%, 20\%, 25\%, 30\%, 35\%, 40\% of the full training set are adopted as the evaluation metric. Every method is run 5 times and the average mIoUs are reported. 

\textcolor{black}{Refer to Table~\ref{tab:iou} for the per-class IoU and mIoU for each method at the fifth AL cycle, using 40\% training data in the Cityscapes dataset. Compared to other popular approaches such as DEAL\cite{Xie_2020_ACCV} and Core-set\cite{sener2017active}, SegXAL is found to be outperforming in overall mIoU (Pseudolabels-63.56; Manual-M -64.37; Manual-D -65.11)}, as well as on various classes, such as road, building, wall, traffic light, traffic sign, vegetation, terrain, sky, rider, car and truck. Furthermore, between the two modes of oracle annotation i.e. Pseudolabel vs Manual, we observe that the manual mode outperforms with a 0.8\% increase against the former, and has a significant boost in class-wise IoUs. \textcolor{black}{We also provide a statistical measure of standard deviation (STD) to give an insight into the variability of the model performance. Further, Table~\ref{tab:miouTrend} displays the incremental trend of mIoU values over multiple iterations. It is observed that at the end of 5 AL cycles itself, mIoU is improved from 20.71 to 63.56 using Pseudolabels, 23.62 to 64.37 using Manual-M and 24.24 to 65.11 using Manual-D.}

\begin{table}[t!]
 \scriptsize

 \caption{\small{Class-wise IoU and mIoU on Cityscape dataset with 40\% training data. For clarity, only the average of 5 AL runs are reported, and the best and the second best results are highlighted in \textbf{bold} and \textit{italics}. }}

\vspace{-.5cm}
 \begin{center}
\setlength{\tabcolsep}{0.1pt} % Adjust column separation for better fit
\renewcommand{\arraystretch}{1.2}
\begin{tabular}{cccccccccccc} 
 \hline
 \hline
\textbf{Method} & \textbf{Road} &  \textbf{Sidewalk} & \textbf{Building} &  \textbf{wall} & \textbf{Fence}& \textbf{Pole} &  \textbf{Traffic Light} & \textbf{Traffic  sign} &  \textbf{Vegetation} & \textbf{Terrain}\\

 \hline
 Fully-supervised  &  \textbf{97.58} & \textbf{80.55}&  \textbf{88.43}&  \textbf{51.22}&  \textbf{47.61}&  35.19 &  \textbf{42.19}&  \textbf{56.79}& \textbf{89.41}&\textbf{60.22}\\

Random~\cite{rangnekar2023semantic}   &  96.03 &  72.36&  86.79&  43.56&  44.22&  36.99 &  35.28&  53.87& 86.91&54.58\\

Entropy~\cite{rangnekar2023semantic}  &  96.28 &  73.31&  87.13&  43.82&  43.87&  \textit{38.10} &  37.74&  55.39& 87.52&53.68\\

Core-Set\cite{sener2017active}  &  96.12 &72.76&  87.03 &44.86&  \textit{45.86}& 35.84 & 34.81 &53.07& 87.18& 53.49\\

DEAL~\cite{Xie_2020_ACCV}  &  95.89 &  71.69&  87.09&  45.61&  44.94&  \textbf{38.29} &  36.51&  55.47& 87.53&56.90\\

\color{black} \textbf{Ours (Pseudolabels)}   &  96.67 &  72.42&  87.04&  \textit{46.91}&  45.02&  36.26&  \textit{37.83}&  56.11& \textit{87.93}& 57.54\\

\color{black} \textbf{Ours (Manual-M)} &   96.91&  72.68&  87.44&  46.62&  45.22 &  35.62&  36.24&  \textit{55.78}& 87.66& 57.86\\

\color{black}\textbf{Ours (Manual-D)} &   \textit{96.98}&  \textit{73.43}& \textit{88.34}&  46.88&  45.38 &  36.12&  37.36& 55.38& 87.84& \textit{59.87}\\

\hline
\hline
\textbf{Method}& \textbf{Sky} &  \textbf{Pedestrian} & \textbf{Rider} &  \textbf{Car} & \textbf{Truck}& \textbf{Bus} &  \textbf{Train} & \textbf{Motor Cycle} &  \textbf{Bicycle} & \textbf{mIoU} & \color{black}\textbf{STD}\\
 \hline

 Fully-supervised  &  \textit{92.69} &  \textbf{65.12}&  37.32&  \textbf{90.67}&  \textbf{66.24}&  \textbf{71.84} &  \textbf{63.84}&  \textbf{42.35}& \textbf{61.84}&\textbf{65.30} & 19.48\\

Random~\cite{rangnekar2023semantic}   &  91.47 &  62.74&  37.51&  88.05&  56.64&  61.00 &  43.69&  30.58& 55.67&59.00 & 20.61\\

Entropy~\cite{rangnekar2023semantic}  &  92.05 &  63.96&  34.44&  88.38&  59.38&  64.64 &  \textit{50.80}&  36.13& \textit{57.10}& 61.46 & 20.14\\

Core-Set\cite{sener2017active}  &  91.89 &62.48& 36.28 &87.63&  57.25 &\textit{67.02}&  56.59 &29.34& 53.56 &60.69 & 20.61\\

DEAL~\cite{Xie_2020_ACCV}   &  91.78 &  \textit{64.25}&  \textbf{39.77}&  88.11&  56.87&  64.46 &  50.39&  \textit{38.92}& 56.59&61.64 & 19.41\\

\color{black} \textbf{Ours (Pseudolabels)}  &  92.18 &  62.53&  38.82&  \textit{88.61}&  59.07&  65.72&  47.12& 35.41& 55.83& 63.56& 20.12\\

\color{black} \textbf{Ours (Manual-M)} & 92.84& 62.73& \textit{39.34}& 87.97& 59.43& 66.01& 46.92& 34.98& 54.93& 64.37 & 19.96\\

\color{black}\textbf{Ours (Manual-D)} & \textbf{92.93}& 62.56& \textit{39.07}& 88.11& \textit{59.47}& 65.70& 46.88& 35.53& 54.71& \textit{65.11} & 20.15\\

\hline
 \hline
 \vspace{-.3cm}
\end{tabular}
\label{tab:iou}
\end{center}
\vspace*{-\baselineskip}
\end{table}

\begin{table}[t]
\centering

 \caption{\small{\textcolor{black}{Comparison of mean IoU of SegXAL model over 5 active learning cycles, with 40\% training data, using pseudo labels annotated by machine vs and human annotations using MiDaS (Manual-M) and DINOv2 (Manual-D) variants.}}}
 \begin{center}
\begin{tabular}{|c|c|c|c|c|c|c|} 
 \hline
\textbf{Mode} & \textbf{ALcycle1} &  \textbf{ALcycle2} & \textbf{ALcycle3} &  \textbf{ALcycle4} & \textbf{ALcycle5} \\
 \hline
 \hline
 \textcolor{black}{Pseudolabel}  &  20.71 &  27.11&  39.23&  50.47&  63.56\\
 \hline
 \textcolor{black}{Manual-M}  & 23.62& 28.02&  39.11&  51.33& 64.37\\
 \hline
 \textcolor{black}{Manual-D} & 24.24 & 30.02 & 39.96 & 52.31 & 65.11\\
 \hline
 \hline
\end{tabular}
\label{tab:miouTrend}
\vspace{-.6cm}
\end{center}
\vspace*{-\baselineskip}
\end{table}

\begin{figure}[ht!]
\scriptsize
  \centering
\subfigure[Entropy- AL cycle1]{\includegraphics[width=2cm]{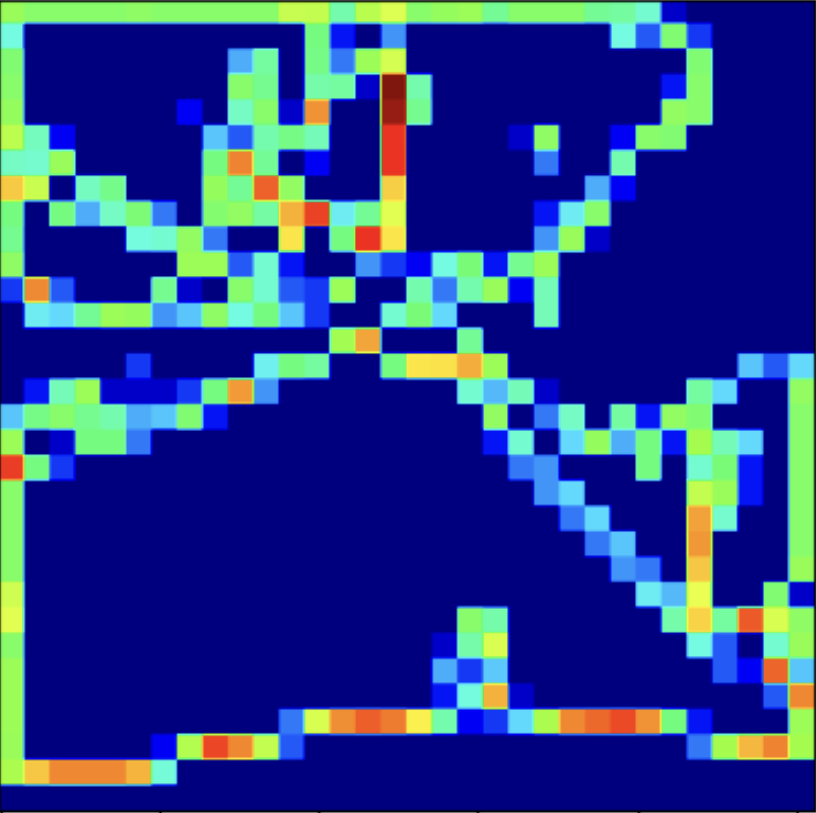}}
\subfigure[Entropy- AL cycle2]{\includegraphics[width=2cm]{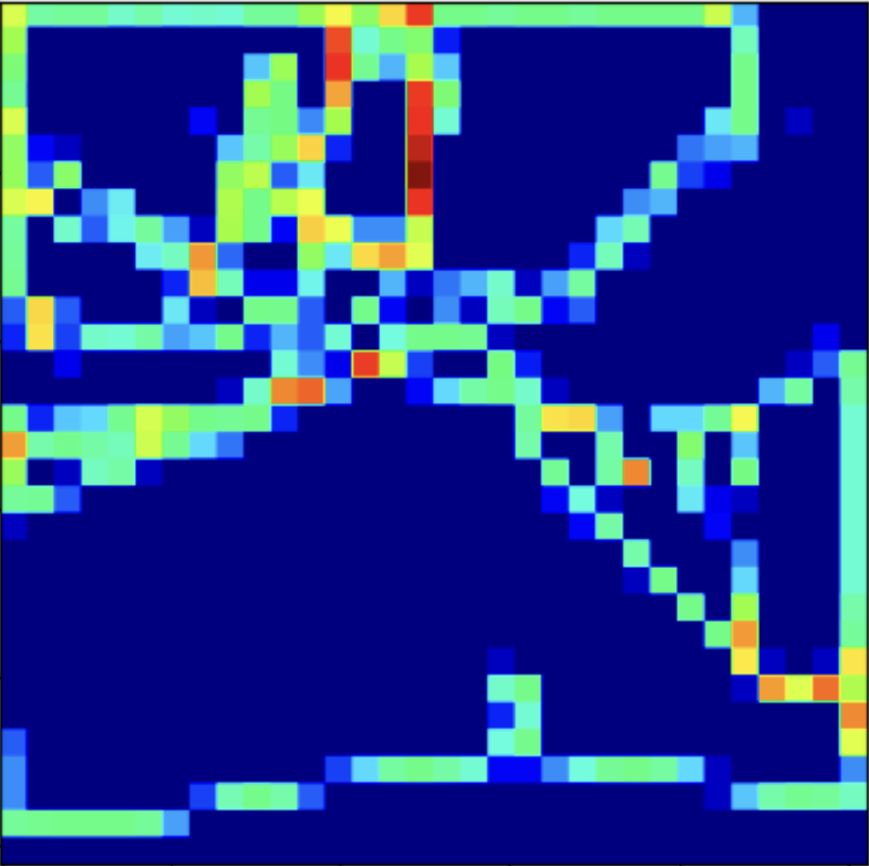}} 
\subfigure[Entropy- AL cycle3]{\includegraphics[width=2cm, height=2cm]{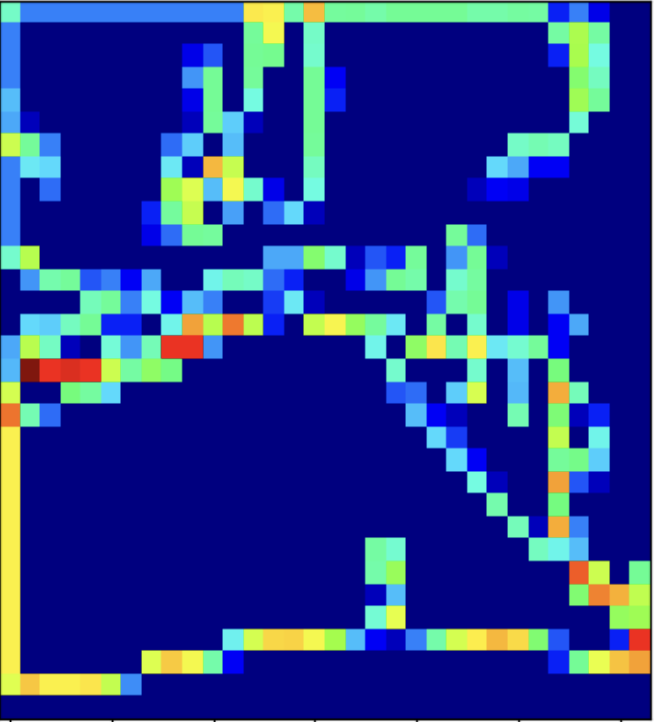}} 
\subfigure[Entropy- AL cycle4]{\includegraphics[width=2cm]{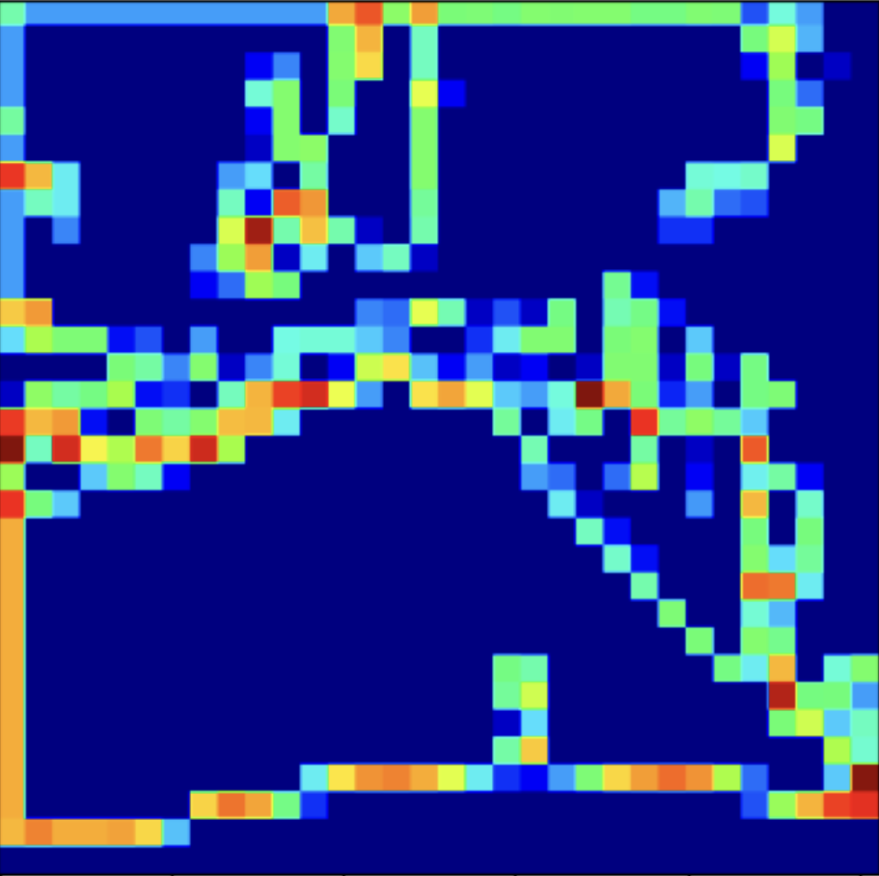}} 
\subfigure[Entropy- AL cycle5]{\includegraphics[width=2cm]{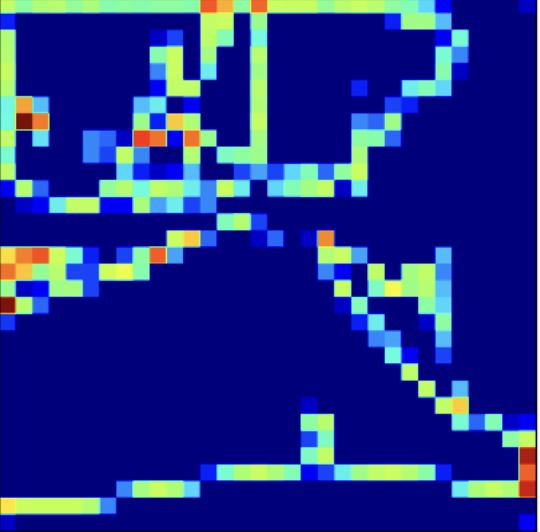}} 

\subfigure[EEM -AL cycle1]{\includegraphics[width=2cm]{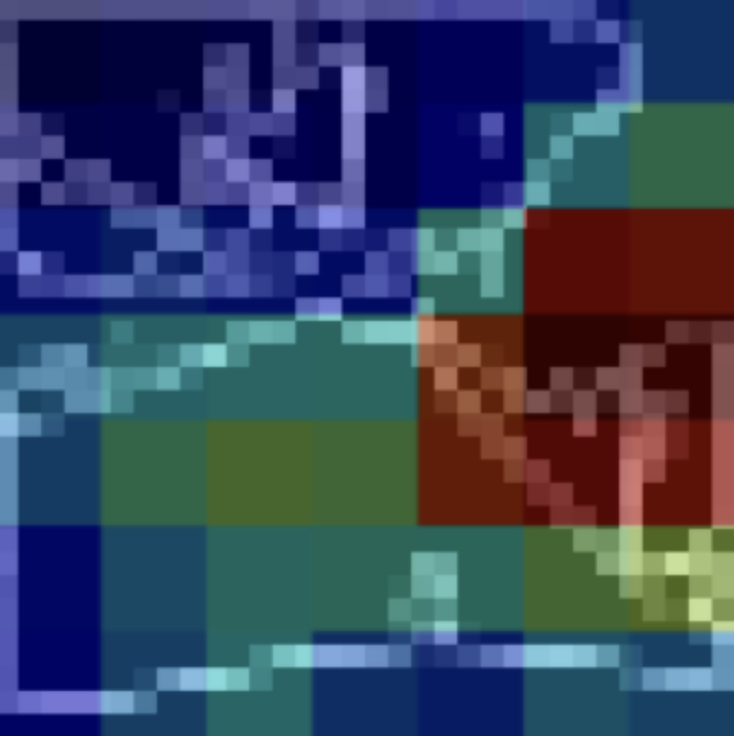}}
\subfigure[EEM -AL cycle2]{\includegraphics[width=2cm]{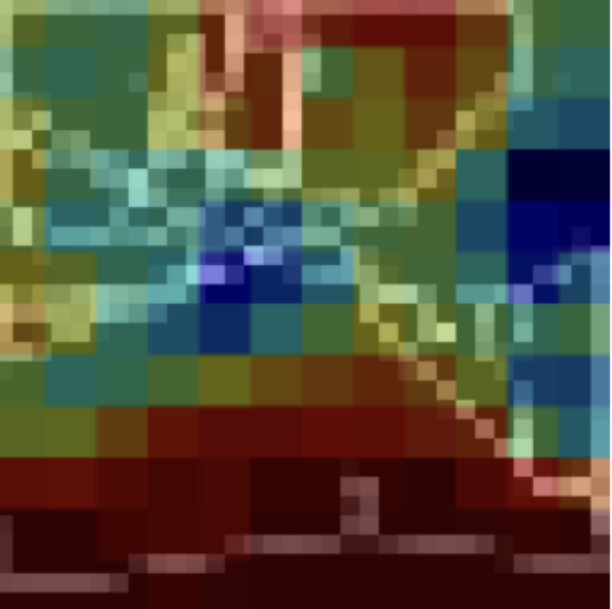}} 
\subfigure[EEM -AL cycle3]{\includegraphics[width=2cm, height=2cm]{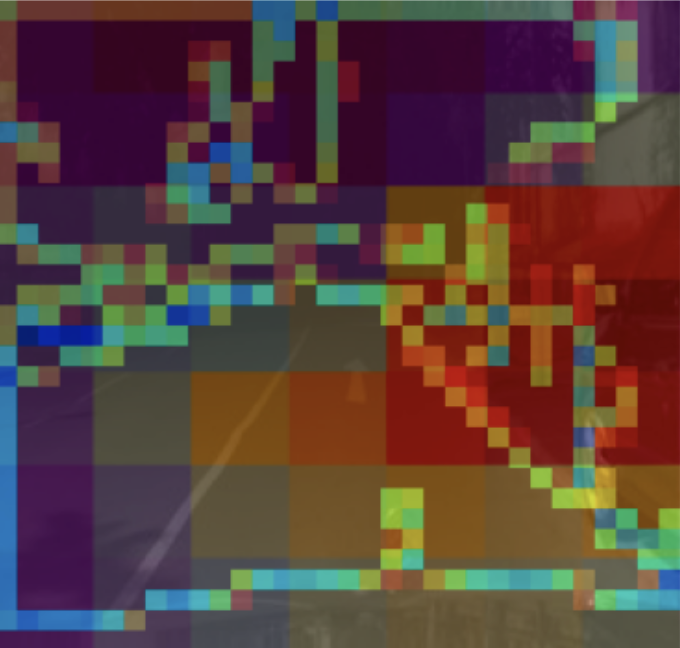}} 
\subfigure[EEM -AL cycle4]{\includegraphics[width=2cm]{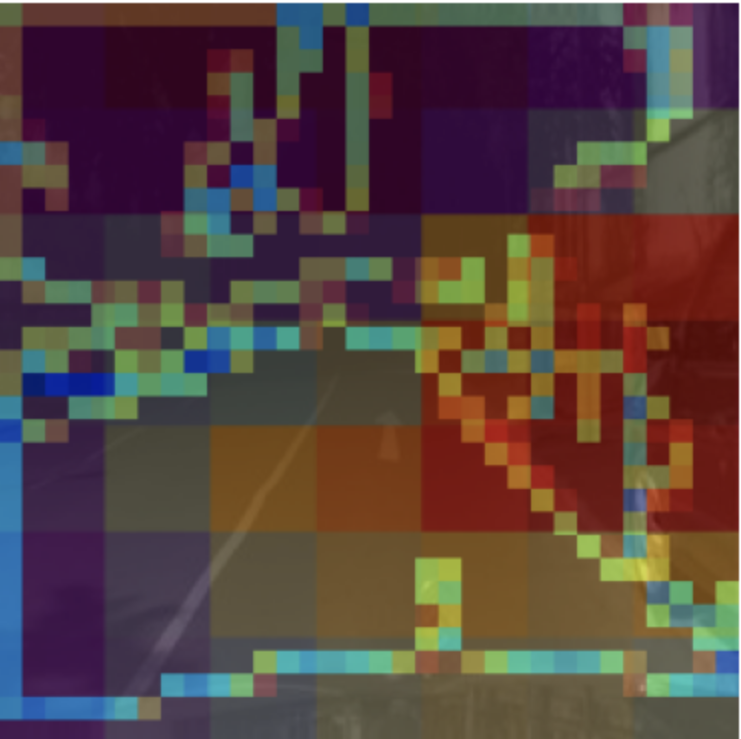}} 
\subfigure[EEM -AL cycle5]{\includegraphics[width=2cm]{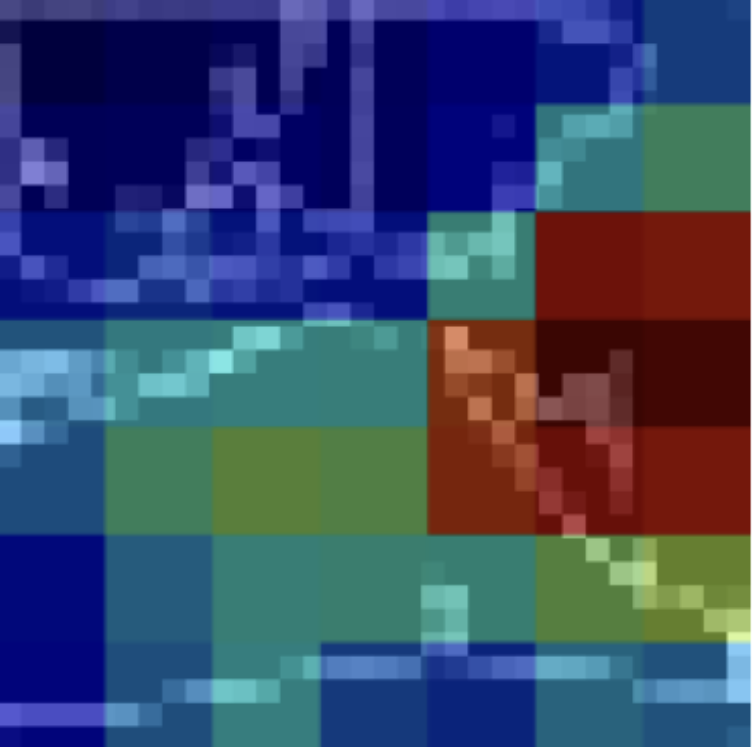}} 

\subfigure[Segmentation-AL cycle1]{\includegraphics[width=2cm]{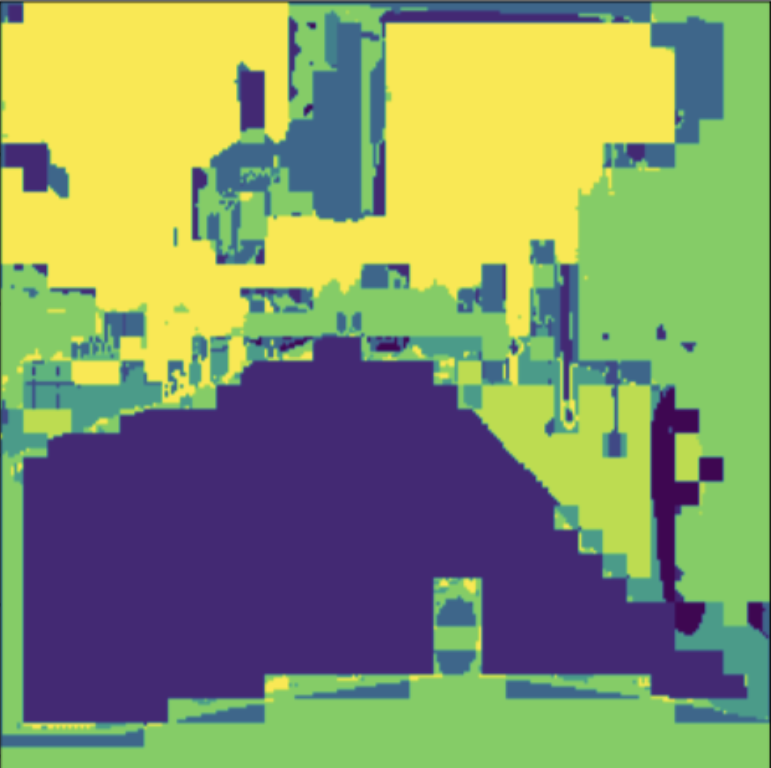}}
\subfigure[Segmentation-AL cycle2]{\includegraphics[width=2cm]{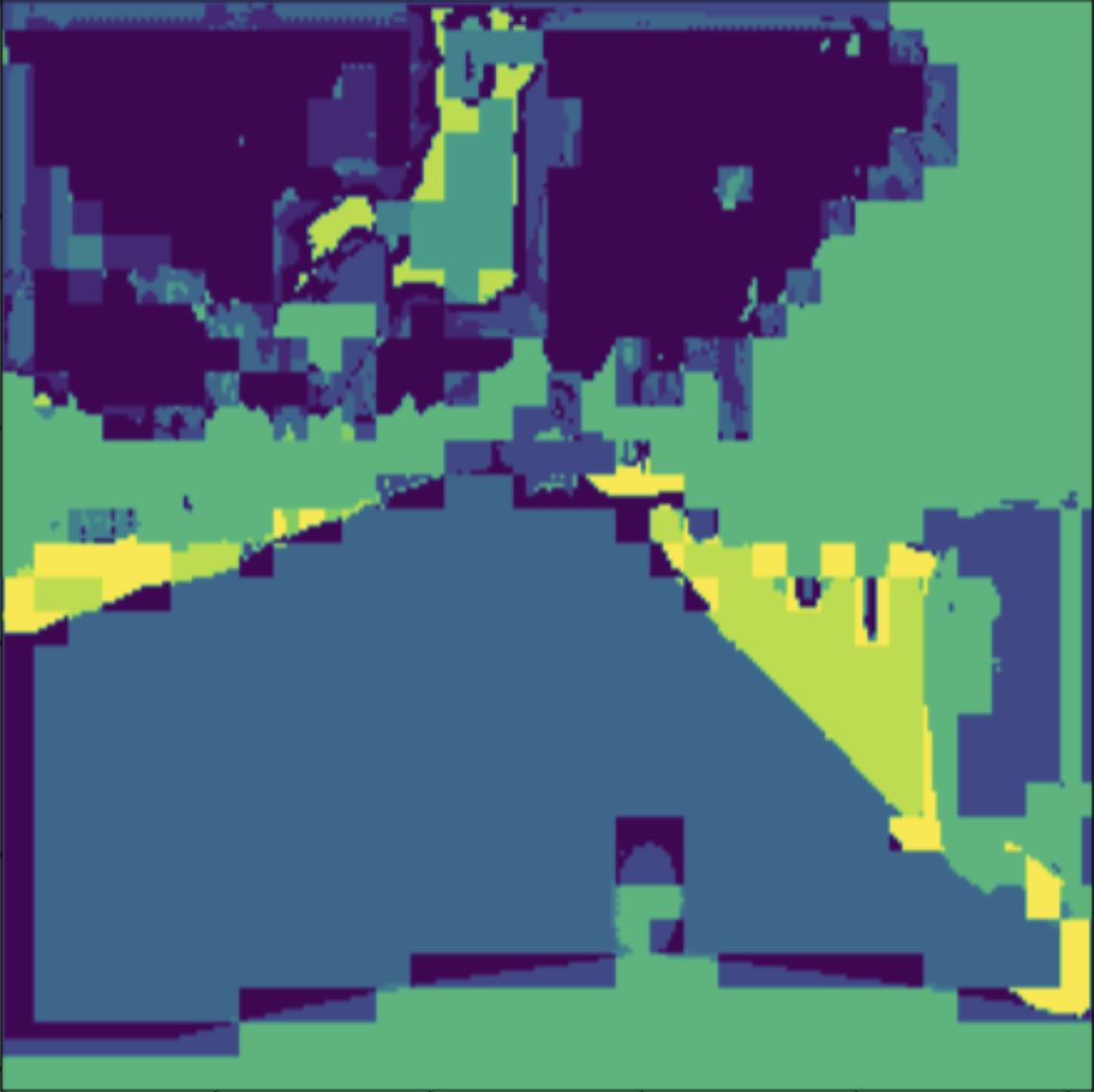}} 
\subfigure[Segmentation-AL cycle3]{\includegraphics[width=2cm]{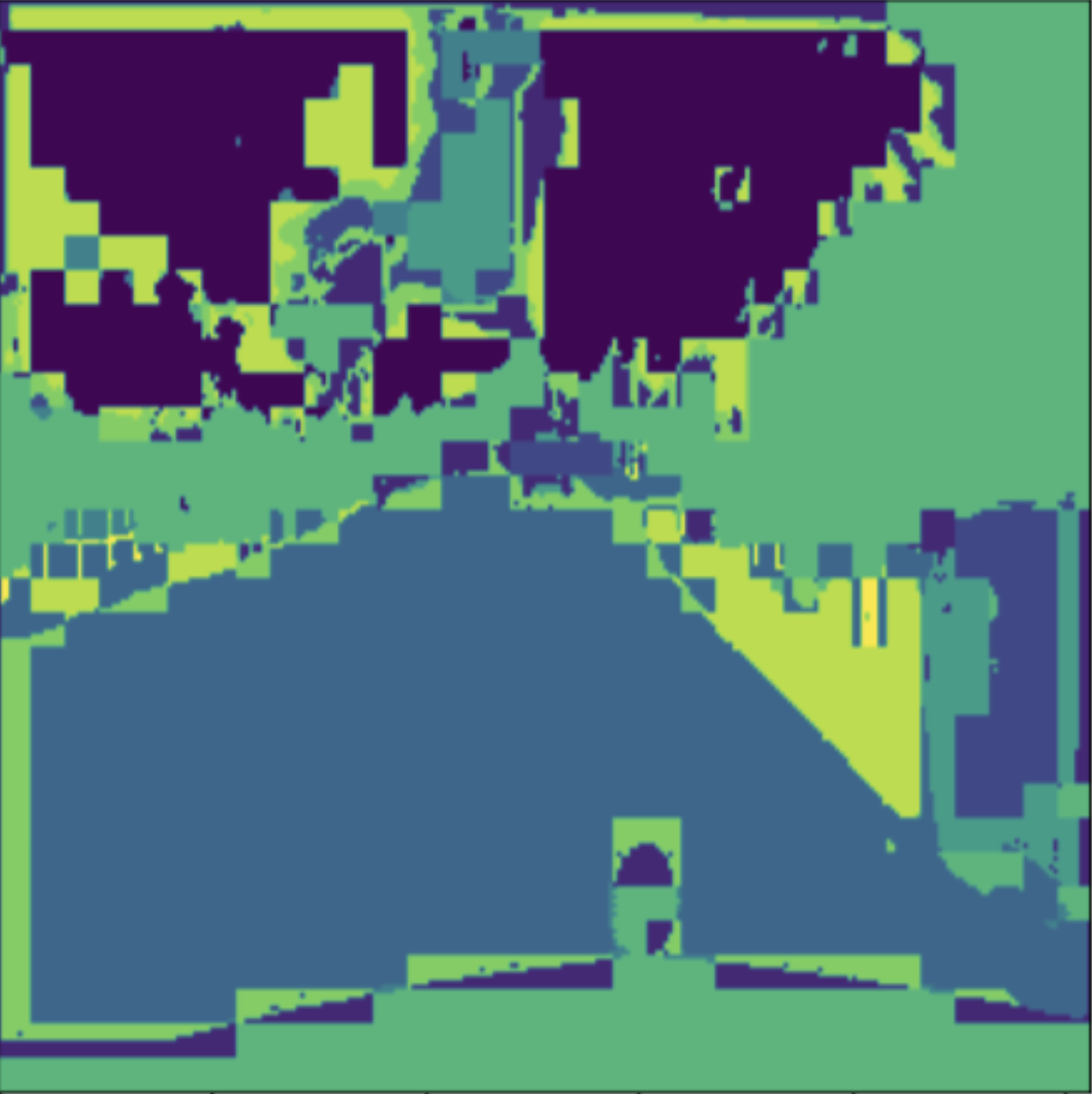}} 
\subfigure[Segmentation-AL cycle4]{\includegraphics[width=2cm]{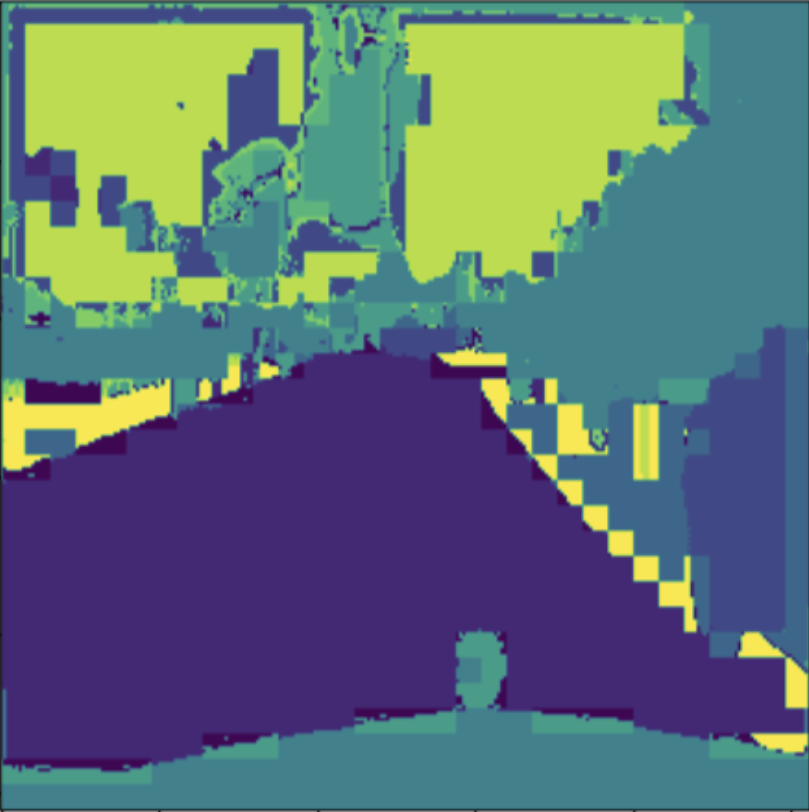}} 
\subfigure[Segmentation-AL cycle5]{\includegraphics[width=2cm]{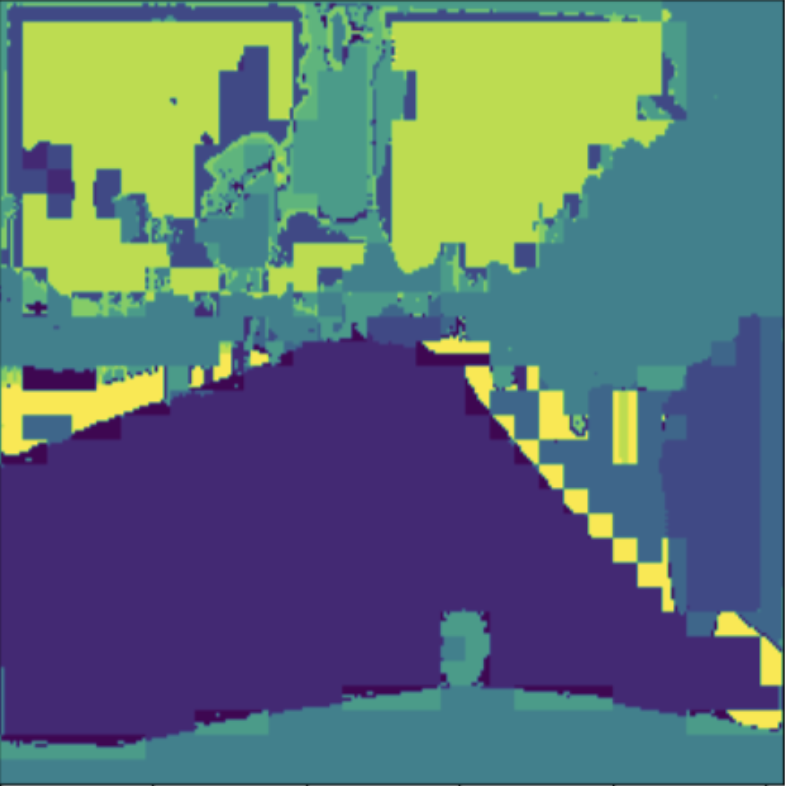}} 

\vspace{-.3cm}
\caption{\small {Visualization of model performance over 5 Active Learning cycles.}}
\vspace{-.5cm} 
\label{fig:entropy-eem}
\end{figure}

\vspace{-.6cm}
\subsection{Visualisation results}
\vspace{-.3cm}
To demonstrate the efficacy of our proposed EEM module,  we visualize the qualitative results. Referring to Fig \ref{fig:entropy-eem}, the visualization of 5 AL cycles of a sample raw image shown in Fig.~\ref{fig:PAE} are depicted column-wise. The pixel entropy, Explainable Error Mask (EEM) output and the machine annotated pseudolabel-based segmentation results are shown along the first, second and third rows respectively. 

Referring to Fig.~\ref{fig:entropy-eem} (a)-(e), high entropy areas represented in red or orange patches indicate a high degree of variability in pixel values. Conversely, low entropy regions, in blue, signify homogenous or less complex segments, where pixel intensities are similar and are in their class boundaries. This entropy map thus serves as a useful visualisation tool to analyse the complexity of the scenes over the loops. Further, Proximity-aware GradCAM-XAI is fused with this entropy mask to obtain an Explainable Error Mask  as shown in Fig.~\ref{fig:entropy-eem} (f)-(j)(Refer Sec. 3.2). These EEM outputs clearly ``\textit{explain}" the oracle to focus and prioritise the annotation of the closer objects/regions with \textit{high entropy},  which are quite critical in decision-making in the real-world scenario. The oracle-annotated results Fig.~\ref{fig:entropy-eem} (k)-(o)depicts the significant improvement in segmentation quality over 5 active learning cycles.

\vspace{-.6cm}
\subsection{Ablation Study}

\subsubsection{i) Impact of \textcolor{black}{Machine based pseudo label annotation} vs Manual Annotation/ Impact of Pixel level strategy and object level strategy:\\}

\begin{figure}[h]
\scriptsize
  \centering
\subfigure[Sample Raw Image]{\includegraphics[width=2.2cm, height=2cm]{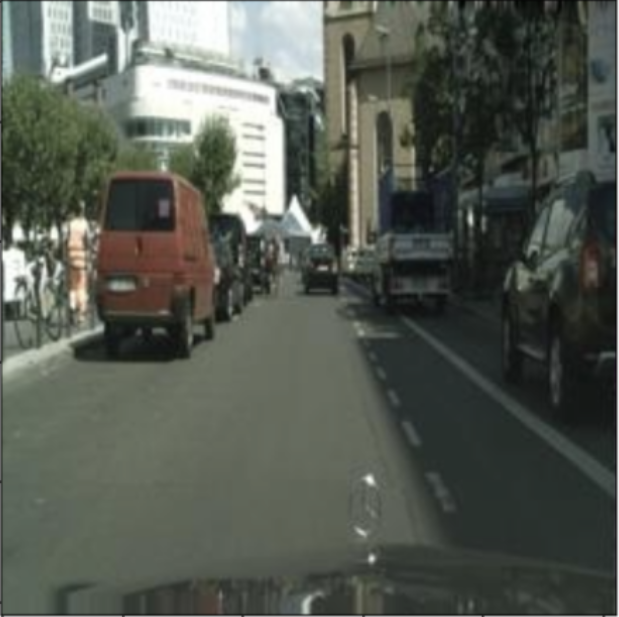}}
\subfigure[Initial Segmentation]{\includegraphics[width=2.2cm, height=2cm]{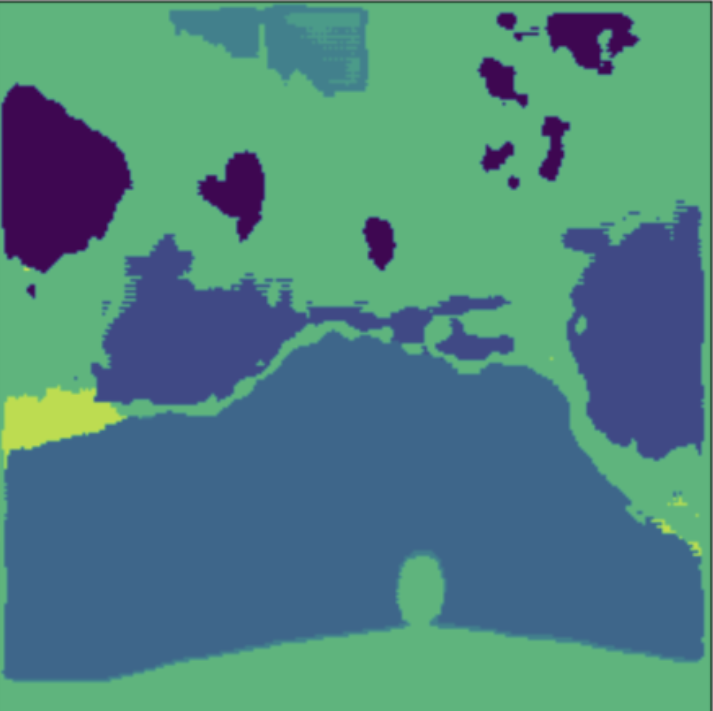}}
\subfigure[Pseudolabel@AL1]{\includegraphics[width=2.2cm, height=2cm]{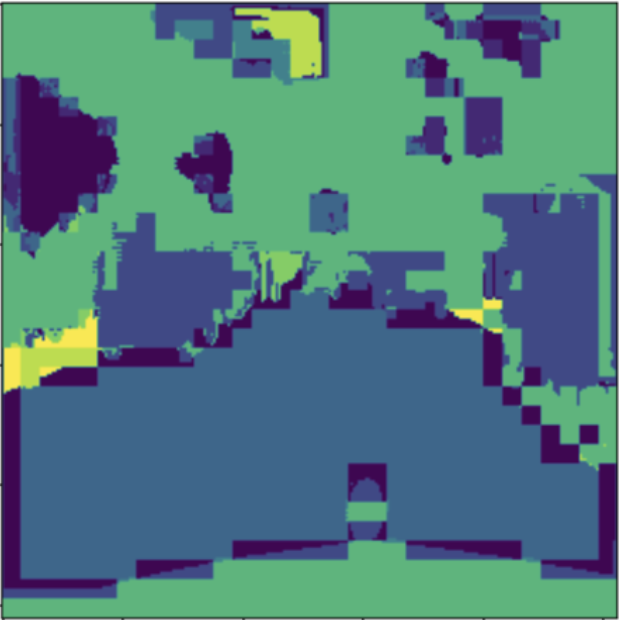}} 
\subfigure[Manual annotation@AL1]{\includegraphics[width=2.2cm, height=2cm]{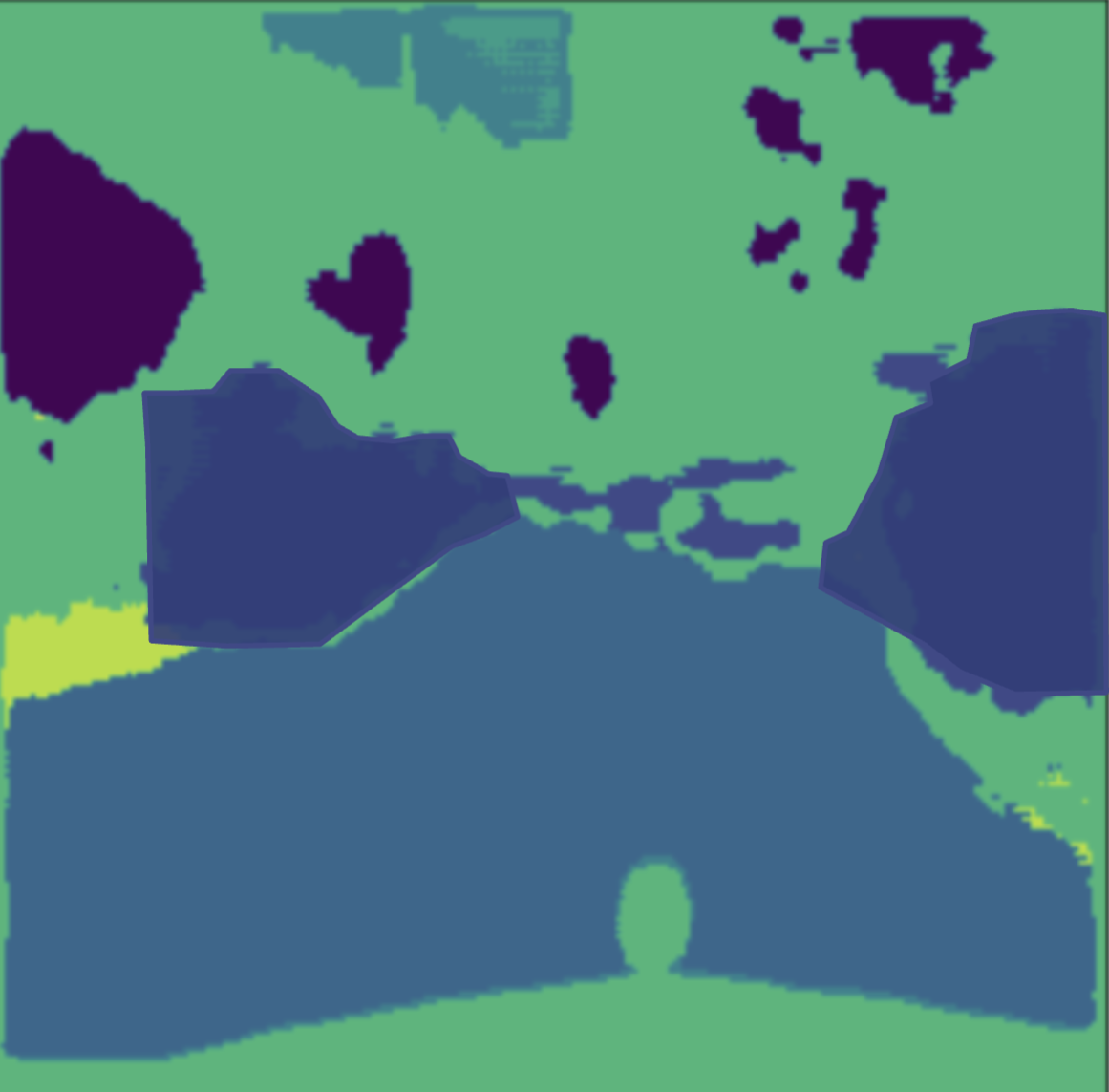}}
\vspace{-.4cm}
\caption{\small Visualization of Machine-based Pseudolabel  vs. Manual annotation outputs}
\label{fig:Sample Oracle}
\end{figure}

In this ablation study, we analyse the effect of \textcolor{black}{Machine-based pseudo label reannotation }and Manual reannotation. As mentioned earlier, the machine oracle mode leverages pixel-level pseudolabel values for annotation whereas the human oracle employs object-level annotation via Label Studio. We could observe from Table~\ref{tab:iou}, Table~\ref{tab:miouTrend} and Fig.~\ref{fig:Sample Oracle} that both approaches provide superior performance in semantic segmentation. Specifically, the manual annotation outperforms the Pseudolabel annotation (Refer Table~\ref{tab:iou}) and smooth segmentation masks (See Fig.~\ref{fig:Sample Oracle}). Nevertheless, Machine-based auto labelling is faster and bestows a promising automated AL solution from a practical perspective compared to manual annotation, wherein a human expert reviews every image and reannotates.\\

\vspace{-.6cm}

\vspace{-.6cm}
\begin{figure}[h!]
  \centering

\subfigure[Image]{\includegraphics[width=2.5cm, height=2cm]{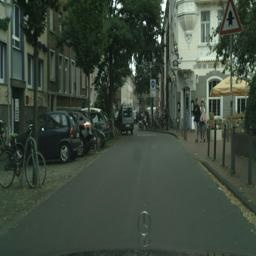}}
\subfigure[GradCAM]{\includegraphics[width=2.5cm, height=2cm]{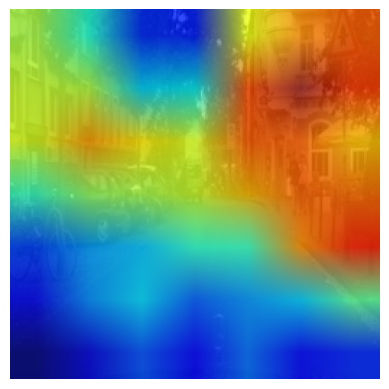}}
\subfigure[PAE]{\includegraphics[width=2.5cm, height=2cm]{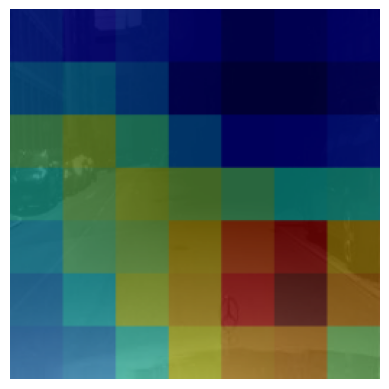}}
\vspace{-.5cm}
\caption{\small Visual representation of Proximity-aware XAI (PAE)}

\vspace{-.5cm}
\label{fig:XAI}
\end{figure}

\subsubsection{ii)Impact of Proximity-aware XAI EBU and PAE modules}:\\
\textcolor{black}{To understand the impact of EBU and PAE modules, quantitative ablation studies are carried out. Referring to Table \ref{tab:miouTrend-ablation}, it can be observed that the lack of EBU sub-module within the EEM block results in a mIoU drop of 3.69, 3.94 and 4.02 in Pseudolabel, Manual-M and Manual-D cases, respectively. Its counterpart results in the absence of PAE sub-modules are 3.47, 2.69 and 3.4 respectively.} Additionally, a qualitative study is also conducted to comprehend the visual interpretation of Proximity-aware XAI, as depicted in Fig.\ref{fig:XAI}.  It is observed that PAE outperforms the Vanilla GradCAM\cite{selvaraju2017grad}, which provides insights of the scene by localizing on the key areas semantic classes via saliency heat maps (Refer Fig.\ref{fig:XAI}(a, b)). Built on top of this Grad-CAM concept, our Proximity-aware XAI module refines the attention further onto the nearby objects in the proximity regions e.g. nearby vehicles and sidewalks, as shown in Fig.\ref{fig:XAI}(c). This PAE enhancement notably fosters safety and transparency in autonomous driving scenarios.

\vspace{-.5cm}
\begin{table}[h]
\centering
 % \vspace*{-\baselineskip}
 \caption{\small{A quantitative study on impact of EEM module and their components.}}
 % \textcolor{red}{What about PseudoLabels? mentioned in the rebuttal.. Rev4, Q1}}}
 \begin{center}
\begin{tabular}{|c|c|c|c|c|c|c|} 
 \hline
\textbf{Mode} & \textbf{\textcolor{black}{PsuedoLabels}} & \textbf{Manual-M} &  \textbf{Manual-D} \\
 \hline
 \hline
 \textcolor{black}{With EEM}  &63.56  &  64.37 &  65.11\\
 \hline
 \textcolor{black}{Without EBU}  &59.87   & 60.43& 61.09\\
 \hline
 \textcolor{black}{Without PAE} &60.09    & 61.68& 61.87\\
 \hline
 \hline
\end{tabular}
\label{tab:miouTrend-ablation}
\end{center}
\vspace*{-\baselineskip}
\end{table}

\vspace{-1.5cm}
\subsubsection{iii)Impact of change in \% of data split}
In this ablation study, we investigate the effect of data split on the SegXAL performance. In particular, we perform various splits of 10\%, 15\%, 20\%, 25\%, 30\%, 35\%, and 40\% of the dataset for the initial model training. Referring to Fig. 7 showing the fifth AL cycle mIoU result, it can be observed that based on the increase of labelled data from 10\% to 40\%, there is a significant increase in mIoU for Pseudolabel 51.02 to 63.56, Manual-M 52.29 to 64.37 and Manual-D 52.83 to 65.11. 

\begin{figure}[h!]
\centering
\includegraphics[width=0.6\textwidth]{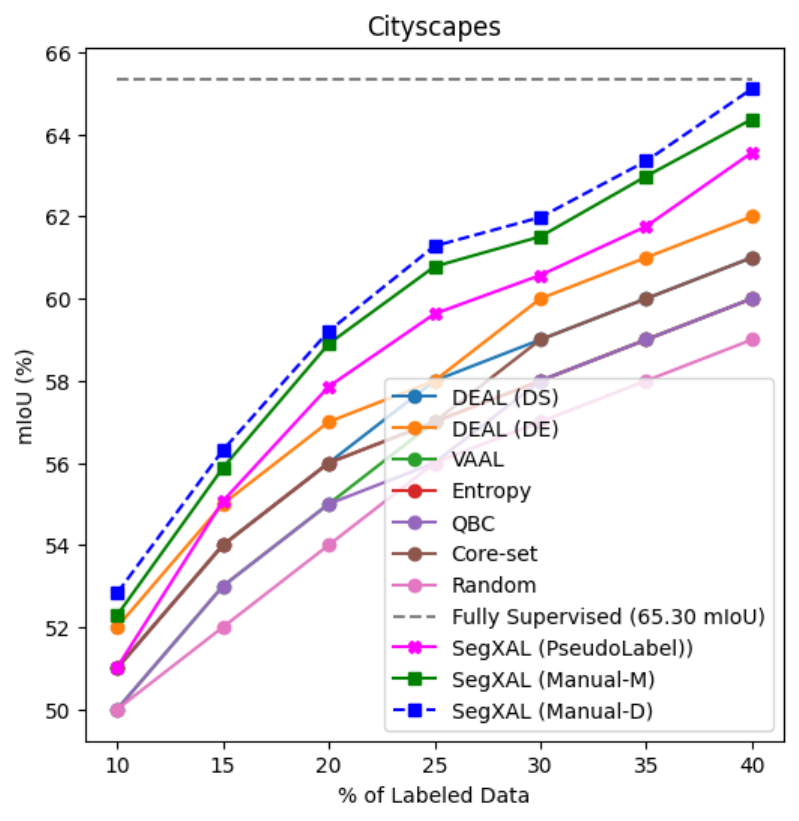}
\textcolor{black}{\caption{SegXAL performance against state-of-the-art on the Cityscapes dataset with 40\% training data.  Every method is evaluated at the end of 5 AL cycles.}}
\label{fig:sotagraph}
\vspace{-.7cm}
\end{figure}

\vspace{-.4cm}
\subsection{State-of-the-art Comparison}
\vspace{-.3cm}
We compare SegXAL with other Active Learning-based semantic segmentation approaches that are deployed on the Cityscapes dataset under similar conditions (with 40\% training data over 5 AL cycles) i.e. DEAL\cite{Xie_2020_ACCV}, core-set approach\cite{sener2017active}, random, entropy\cite{Xie_2020_ACCV,rangnekar2023semantic} and QBC \cite{Xie_2020_ACCV}. {Although another recent study S4AL\cite{rangnekar2023semantic} achieves a competitive result of mIoU 64.80, it is not included in the comparison due to its different setting of 16\% training data.} Referring to the results as shown in Table~\ref{tab:iou} and Fig. 7, it can be observed that SegXAL outperforms the state-of-the-art approaches with a significant margin, achieving the best result of 65.11 mIoU with  human annotations with DINOv2 depth map (blue-dotted line). It is also observed from Table~\ref{tab:iou} that, the segmentation performance on the nearby classes such as road (96.98), sidewalk (73.43), wall (73.48) and vehicles such as truck (59.47), rider (39.34) are better or on par with the previously proposed methods. This superior performance could be accredited to the Explainable Error Mask module that facilitates object-level proximity mechanism using XAI attention and Entropy metric, which prioritizes the highly informative nearby objects' annotations compared to far away objects such as train, vegetation etc.

\vspace{-.6cm}
\section{Conclusions and Future work}
\label{sec:conclusion}
\vspace{-.4cm}
In this work, we proposed a novel Explainable Active Learning framework viz. SegXAL for semantic segmentation. A pilot study on the application of the SegXAL model for driving scene semantic segmentation is presented in this paper. In contrast to most of the existing Active learning methods that annotate using uncertainty information, the proposed model additionally ``explains" the proximity region of interests and key objects to be prioritized while annotating by the oracle, with the help of a newly proposed Explainable Error Mask (EEM) module. Such XAI heatmap explanations not only improve the segmentation accuracy but also bridge the semantic gap that exists between human and machine interpretation. Our SegXAL model outperforms state-of-the-art results. Future improvements can be made by introducing better attention mechanisms such as Vision transformers and extending the applications to other driving datasets and other domains.

\vspace{-.4cm}

\bibliographystyle{IEEEtran}
\bibliography{references}

\end{document}